\pdfoutput=1
\documentclass[11pt]{article}

\usepackage{style/emnlp2025}

\usepackage{times}
\usepackage{latexsym}

\usepackage[T1]{fontenc}

\usepackage[utf8]{inputenc}

\usepackage{microtype}

\title{Leveraging In-Context Learning for Language Model Agents}

\author{
Shivanshu Gupta$^{1}$\thanks{Work done while authors were at Allen Institute for AI.} ~~~ Sameer Singh$^{1}$ ~~~ Ashish Sabharwal$^{2}$ ~~~ Tushar Khot\footnotemark[1] ~~~ Ben Bogin\footnotemark[1] \\
$^{1}$University of California Irvine \hspace{4mm} $^{2}$Allen Institute for AI  \hspace{4mm} \\
\texttt{\makecell{\{shivag5,sameer\}@uci.edu, ashishs@allenai.com}}
}

\usepackage{graphicx}
\usepackage{amssymb}
\usepackage{amsmath}
\usepackage{breqn}
\usepackage{booktabs}
\usepackage{makecell}
\usepackage{multicol}
\usepackage{multirow}
\usepackage{enumitem,kantlipsum}
\usepackage{hyperref}
\usepackage{algorithm,algpseudocode}
\usepackage{arydshln}
\usepackage{caption}
\usepackage{bbm}
\usepackage{soul}
\usepackage{subcaption}
\usepackage{inconsolata}
\usepackage{tabularx}
\usepackage{import}
\usepackage{listings}
\usepackage{tabularray}
\usepackage{amsthm}

\newsavebox\mybox

\theoremstyle{definition}

\begin{document}

\maketitle

\newif\ifdebug

\newif\ifcomments
\commentstrue
\ifcomments
    \providecommand{\sg}[1]{{\protect\color{cyan}{[\textbf{shiv}: #1]}}}
    \providecommand{\ben}[1]{{\protect\color{red}{[\textbf{ben}: #1]}}}
    \providecommand{\sameer}[1]{{\protect\color{brown}{[\textbf{sameer}: #1]}}}
    \providecommand{\ashish}[1]{{\protect\color{brown}{[\textbf{ashish}: #1]}}}
    \providecommand{\tushar}[1]{{\protect\color{yellow}{[\textbf{ben}: #1]}}}
\else
    \providecommand{\sg}[1]{}
    \providecommand{\ben}[1]{}
    \providecommand{\sameer}[1]{}
    \providecommand{\ashish}[1]{}
    \providecommand{\tushar}[1]{}
\fi

\newcommand{\tightparagraph}[1]{\smallbreak\noindent\textbf{#1}}

\definecolor{applegreen}{rgb}{0.01, 0.65, 0.01}
\definecolor{cardinal}{rgb}{0.77, 0.12, 0.23}

\newcommand{\COS}{cosine similarity}
\newcommand{\BSR}{BERTScore-Recall}
\newcommand{\rsc}{\textsc{Random}}
\newcommand{\cossc}{\textsc{Cosine}}
\newcommand{\bsrsc}{\textsc{BSR}}
\newcommand{\setcossc}{\textsc{Set-Cosine}}
\newcommand{\setbsrsc}{\textsc{Set-BSR}}

\newcommand{\neo}{GPT-Neo-2.7B}
\newcommand{\neoemph}{\textbf{\neo}}
\newcommand{\llama}{LLaMA}
\newcommand{\llamaemph}{\textbf{\llama}}
\newcommand{\llamaseven}{LLaMA-7B}
\newcommand{\llamasevenemph}{\textbf{\llamaseven}}
\newcommand{\llamathirteen}{LLaMA-13B}
\newcommand{\llamathirteenemph}{\textbf{\llamathirteen}}
\newcommand{\starcoder}{StarCoder}
\newcommand{\starcoderemph}{\textbf{\starcoder}}
\newcommand{\turbo}{GPT-3.5-Turbo}
\newcommand{\turboemph}{\textbf{\turbo}}
\newcommand{\codex}{Codex}
\newcommand{\codexemph}{\textbf{\codex}}
\newcommand{\cushman}{Cushman}
\newcommand{\cushmanemph}{\textbf{\cushman}}
\newcommand{\davinci}{Davinci}
\newcommand{\davincinemph}{\textbf{\davincin}}

\newcommand{\score}[0]{\mathtt{score}}

\newenvironment{resultstable}[1][]{
    \begingroup
    \setlength{\tabcolsep}{3pt} %
    \begin{table*}[#1]
    \centering
    \small
}{
    \end{table*}
    \endgroup
}

\newenvironment{resultstablesinglecol}[1][]{
    \begingroup
    \setlength{\tabcolsep}{3pt} %
    \begin{table}[#1]
    \centering
    \small
}{
    \end{table}
    \endgroup
}

\begin{abstract}
In-Context Learning (ICL) with dynamically selected demonstrations combines the flexibility of prompting large language models (LLMs) with the ability to leverage training data to improve performance. While ICL has been highly successful for prediction and generation tasks, leveraging it for agentic tasks that require sequential decision making is challenging---one must think not only about how to annotate long trajectories at scale and how to select demonstrations, but also what constitutes demonstrations, and when and where to show them.
To address this, we first propose an algorithm that leverages an LLM with retries and demonstration selection to automatically and efficiently annotate agentic tasks with solution trajectories.
We then show that set-selection of trajectories of similar tasks as demonstrations significantly improves performance, reliability, robustness, and efficiency of LLM agents.
However, trajectory demonstrations have a large inference cost overhead. We show that this can be mitigated by using small trajectory snippets at every step instead of an additional trajectory. We find that demonstrations obtained from larger models (in the annotation phase) also improve smaller models, and that ICL agents can even rival costlier trained agents.
Thus, our results reveal that ICL, with careful use, can be very powerful for agentic tasks as well.
\end{abstract}

\section{Introduction}
\label{sec:intro}

\begin{figure}[ht]
    \centering
    \includegraphics[width=\linewidth]{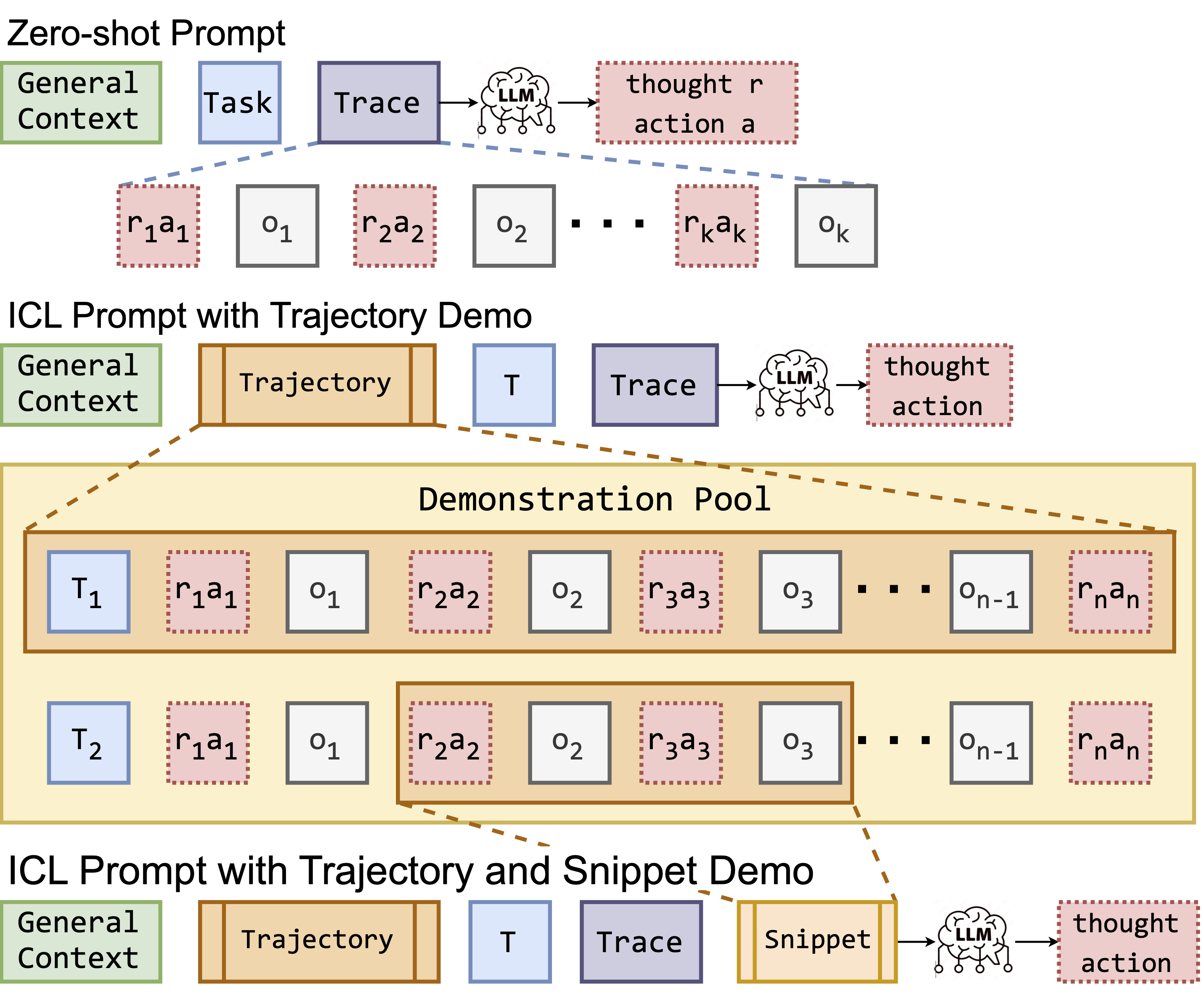}
    \caption{Different types of demonstrations for agentic tasks. \textbf{Top} Using LLMs to simulate agentic behavior involves repeatedly prompting it with the general setup, a task description, and an execution trace recording the agent's thoughts ($r$), actions ($a$), and observations ($o$). \textbf{Middle} Given a pool of tasks paired with solution trajectories, one way to show demonstrations is to use entire trajectories for similar tasks in the prompt. While effective, this has a large overhead. \textbf{Bottom} Another way is to use smaller trajectory snippets with similar reasoning that are post-fixed to the prompt.
    }
    \label{fig:main}
\end{figure}

Due to advances in pretraining, instruction tuning, and scaling, Large Language Models (LLMs) are now increasingly powering autonomous agents to perform complex real-world tasks that require acting in an environment and sequential decision-making. Using LLMs to simulate such agentic behavior involves repeatedly prompting and asking them to generate the next action to be executed. However, LLM agents can be unreliable, especially for complex tasks with long trajectories. Prior work on enhancing LLM agents through structured prompting-based workflows with explicit reasoning, planning, or reflection steps \cite{shinn2023reflexion,kim2023language,sun-etal-2024-pearl} used a fixed prompt for every task instance, without leveraging training data. On the other hand, approaches based on task-specific supervised finetuning or reinforcement learning \cite{chen2023fireact,mitra2024agentinstruct,loop} are too expensive to apply to larger, more powerful LLMs and to update with the knowledge needed for new tasks after training.

In this work, we explore an alternative approach that is prompting-based yet takes advantage of training data, namely \emph{In-Context Learning (ICL) with Demonstration Selection}, where demonstrations relevant to each instance are selected at inference time from a pool of annotations. While ICL is already effective \citep{brown-etal-2020-fewshot}, demonstration selection can dramatically boost it for traditional NLP tasks \cite{gistscore,gupta-etal-2023-coverage,ye2023compositional}. However, unlike such non-sequential tasks where demonstrations can simply be input-output pairs from a train set, for agentic tasks, the training sets of tasks, even when available, are rarely annotated with \textit{solutions} that can serve as demonstrations. Moreover, due to context length limits and recency bias of LLMs, one needs to think not just of \emph{how} to select demonstrations, but also \emph{what} these demonstrations comprise (e.g., entire trajectories or snippets thereof), \emph{when} to show them (i.e., what step), and \emph{where} to place them in the prompt.

To address these challenges, we propose an iterative annotation algorithm that leverages ICL and demonstration selection itself to automatically and efficiently annotate training tasks with solutions that can be used as demonstrations. We then use these annotations to study different demonstration granularities and placements. We begin with the simplest approach, which is to show entire trajectories of similar tasks. As shown in Fig. \ref{fig:main} (middle), these are placed before the test task and shown at every step. Since trajectories tend to be long and only a limited number of them can fit in the prompt, we explore how to select optimal \textit{sets} of trajectories. Finally, as trajectory demonstrations can have a large overhead in terms of inference costs, we explore two forms of smaller demonstrations.
First, we consider smaller subtask trajectories by switching to a Plan-and-Execute (PnE) solver \cite{yang2023intercode,sun-etal-2024-pearl} that decomposes tasks into a sequence of subtasks and executes them one by one. Second, we show small relevant snippets of trajectories at every step (Fig. \ref{fig:main} (bottom)). These are selected based on the agent's reasoning in the latest step and shown at the end of the prompt for a single step, thus accounting for LLMs' recency bias and also having a minimal overhead.

For our testbed, we use the AppWorld benchmark \cite{trivedi-etal-2024-appworld}, which evaluates an LLM agent's ability to carry out complex day-to-day user tasks involving sending emails, making payments, playing music, shopping, etc., by interacting with a variety of apps via their APIs. AppWorld's rich code-based action space, varying observation sizes, and complex tasks make it an ideal testbed for studying various design decisions relating to demonstration selection. Using our annotation algorithm, we automatically annotate over 95\% of training tasks with solutions to create a demonstration pool.
We show that using the annotated trajectory of even a single most similar task as a demonstration boosts a zero-shot agent's performance by 29 points, 16 points more than using a fixed, manually written trajectory. %
Further, when additional trajectories can be included, selecting them jointly as a set \cite{gupta-etal-2023-coverage} is more effective than independent ranking-based selection.

Selecting trajectory demonstrations is particularly effective at improving reliability (across multiple runs of the same task) and robustness (across variations of the same tasks), outperforming the use of a fixed trajectory by 20.8 and 23.2 points, respectively.
However, while effective, trajectory demonstrations also have a large overhead in terms of inference costs---each additional trajectory adding on the order of 100K tokens on average in inference costs.
Instead, as we show, providing small, relevant snippets of the trajectories as demonstrations at every step is also effective at improving performance, notably with negligible overhead. We find a combination of trajectory and snippet demonstrations to be the optimal approach, and with these, a prompted LLM agent can be made competitive with most state-of-the-art training-based approaches \cite{loop}. Overall, our results show that, similar to traditional NLP tasks, demonstration selection can yield significant performance gains for LLM agents and can enable prompted LLM agents to rival even trained ones.

\section{Related Work}
\label{sec:related}

\tightparagraph{LLM Agents.} LLMs are increasingly being used to power autonomous agents for a variety of agentic tasks involving sequential decision-making. These include web navigation to answer user queries \cite{zhou2024webarena,drouin2024workarena} and e-commerce \cite{yao2023webshop}, playing games \cite{shridhar2021alfworld}, interacting with applications and APIs to carry out user tasks \cite{trivedi-etal-2024-appworld}, running ML experiments \cite{bogin2024super}, and more. Prior work on improving agent performance on such tasks has looked into (1) prompting based approaches to inducing structured workflows with explicit reasoning, planning, and reflection steps \cite{react,shinn2023reflexion,wang2024deps,kim2023language,sun-etal-2024-pearl}, (2) training-based approaches including supervised finetuning on agent trajectories and reinforcement learning \cite{nakano2022webgpt,yao2023webshop,deng2023mind2web,chen2023fireact,qin2023toolllm,mitra2024agentinstruct,loop}.

\tightparagraph{In-Context Learning (ICL)} \cite{brown-etal-2020-fewshot} is the ability of LLMs to solve unseen tasks without training by merely conditioning on a few task demonstrations and without any task-specific training. However, ICL performance is highly sensitive to the choice of demonstrations \cite{zhao2021calibrate}, and can be significantly improved by dynamically selecting demonstrations for each test input \cite{liu-etal-2022-makes}.
There is now a large body of work on selecting better demonstrations, exploring among other things, better metrics for scoring demonstration candidates \cite{rubin-etal-2022-learning,gupta-etal-2023-coverage,gistscore,askari2025unraveling}, selecting demonstrations as a set \cite{gupta-etal-2023-coverage,ye2023compositional}, selecting diverse demonstrations to reduce redundancy among them \cite{su2022selective,levy-etal-2023-diverse,agrawal2022incontext,ye2022complementary}, etc.

\tightparagraph{Demonstration Selection for Agentic Tasks.} 
Prior work on demonstration selection for ICL has primarily focused on traditional, non-sequential NLP tasks that involve mapping inputs to outputs. The two prior works that have studied demonstration selection in the context of agentic tasks are Synapse \cite{synapse} and TRAD \cite{trad}. However, they primarily focused on web navigation tasks where the main challenge was the size of individual HTML observations rather than task complexity (in terms of number of steps).
In contrast, we focus on more complex tasks which involve numerous steps with long-range dependencies, but not every step yields a large observation, allowing an entire trajectory or two can fit in the context. This setup allows us to study the impact of different granularities, selection, and placements of demonstrations. Notably, this will also become an increasingly common scenario as LLM context lengths increase, but the cost-benefit trade-offs we explore will remain.

\section{Preliminaries}
\label{sec:prelims}

\subsection{LLM Agents}
\label{sec:agents}

\tightparagraph{ReAct.} The predominant approach to creating LLM-powered agents for agentic tasks is ReAct \cite{react}. As shown in Fig. \ref{fig:main} (Top), it involves repeatedly prompting the LLM with a trace of past execution and asking it to produce a \textit{thought} (denoted $\mathbf{r}$) describing its reasoning about its progress and an \textit{action} (denoted $\mathbf{a}$), to be executed in the environment to obtain the next observation (denoted $\mathbf{o}$).
Formally, given (1) a context $\mathbf{c}=\langle \mathbf{p}, \mathbf{q} \rangle$ comprising a general context $\mathbf{p}$ which describes the setup, provides demonstrations and guidelines, etc., and a task-specific context $\mathbf{q}$ describing the task to be carried out, and (2) a trace of past thoughts, actions, and observations $\mathbf{h}_t=\langle \mathbf{r}_1, \mathbf{a}_1, \mathbf{o}_1, \ldots, \mathbf{r}_t, \mathbf{a}_t, \mathbf{o}_t\rangle$,
the LLM is prompted to generate the next thought and action:
\begin{equation}
    \mathbf{r}_{t+1}, \mathbf{a}_{t+1} \sim \mathcal{P}_{LM}\left(\cdot \mid \mathbf{c}, \mathbf{h}_t\right)
\end{equation}
The next observation $\mathbf{o}_{t+1}$ is then obtained by executing the action $\mathbf{a}_{t+1}$ in the environment. This process is repeated until a terminal state is reached.
We will refer to the complete execution trace $\mathbf{h}_T$ as a trajectory $\tau$.

\tightparagraph{Plan \& Execute (PnE).} The ReAct approach, as described above, tries to solve the entire task in one go and retains the entire execution trace in the prompt. However, this can be expensive as the trajectories for complex agentic tasks are often very long. One way to address this is to use a Plan \& Execute (PnE) approach \cite{yang2023intercode,sun-etal-2024-pearl}. PnE takes advantage of the fact that a task may involve multiple simpler subtasks, and how each subtask is carried out may not be relevant to the other subtasks. It incorporates a planning step that breaks down the original task $\mathbf{t}$ into a sequence of subtasks $\mathbf{t^1}, \ldots, \mathbf{t^m}$. Each subtask is then executed by a ReAct-styled executor agent, optionally with the plan and summaries of previous subtasks' trajectories provided in the task-specific prompt $\mathbf{q}$.

\subsection{In-Context Learning and Demonstration Selection}
\label{sec:icl}

In-Context Learning (ICL) is the ability of LLMs to solve unseen tasks by conditioning on a few task demonstrations. Formally, for traditional NLP tasks, given demonstrations in the form of input-output pairs $\left\{\left(\mathbf{x_i}, \mathbf{y_i}\right)\right\}_{i=1}^k$ and the test input $\mathbf{x}_{\text {test}}$, it involves prompting the LLM with the context $\mathbf{c}=\langle \mathbf{x}_1, \mathbf{y}_1, \ldots, \mathbf{x}_k, \mathbf{y}_k, \mathbf{x}_{\text{test}}\rangle$ and letting it generate the output $\mathbf{y}_{\text{test}}$. Although using the same demonstrations for all test inputs allows ICL to work even for tasks lacking any training data, when a training set $\mathcal{T}=\left\{\left(\mathbf{x}_i, \mathbf{y}_i\right)\right\}_{i=1}^N$ is available, performance can be boosted using some form of demonstration selection \cite{liu-etal-2022-makes,rubin-etal-2022-learning,gupta-etal-2023-coverage,gistscore}. Using the training set as a pool of demonstration candidates, it involves selecting $k \ll N$ demonstrations that, when placed in the context, increase the likelihood of the correct output being generated. Some approaches for demonstration selection proposed for traditional NLP tasks that we will experiment with include:

\tightparagraph{Ranking-based Selection.}
This involves scoring all the candidates for their relevance with respect to the test instance and using the top-K candidates as demonstrations. Formally, given the training set $\mathcal{T}=\left\{\left(\mathbf{x}_i, \mathbf{y}_i\right)\right\}_{i=1}^N$ and the test input $\mathbf{x}_{\text{test}}$, the demonstrations are selected as $\operatorname{topk}_{i} \operatorname{sim}(\mathbf{x}_{\text{test}}, \mathbf{x}_i)$, where $\operatorname{sim}(\cdot)$ is a similarity metric. Note that the metric operates on only the inputs that proxy as the \textit{retrieval key} for the demonstrations. Prior work has explored Cosine Similarity \cite{liu-etal-2022-makes} and BERTScore-Recall (BSR) \cite{zhang2020bertscore,gupta-etal-2023-coverage} as metrics, both of which involve encoding the retrieval key using a dense encoder and using it to identify the closest candidate.

\tightparagraph{Set Selection.} \citet{gupta-etal-2023-coverage} showed that ranking-based selection can be sub-optimal for complex compositional tasks as it may select demonstrations that are individually relevant to the test input yet fail to provide all the relevant information needed to solve it. Instead, they argue that demonstrations should be selected as a set such that they cover all the reasoning patterns. They proposed Set-BSR, a set-extension of BSR, that is submodular and hence greedily optimizable.

\section{Automatic Trajectory Annotation}
\label{sec:annotation}

\begin{algorithm}[tb]
    \small
    \caption{Iterative Annotation for Agentic Tasks} %
    \label{alg:iterative}
    \begin{algorithmic}[1]
        \Require Task pool $\mathcal{T}$; Demonstration selector $D$; Solver $S$; Solution Checker $C$; Number of Rounds $R$
        \State $\mathcal{U} \gets \mathcal{T}$\Comment{Unannotated tasks}
        \State $\mathcal{T^*} \gets \emptyset$\Comment{Annotated tasks}
        \For{$r=1$ to $R$}
            \For{$\mathbf{t} \in \mathcal{U}$}
                \State $\mathcal{D} \gets D(\mathbf{t}, \mathcal{T^*})$\Comment{Select demonstrations}
                \State $\mathbf{s} \gets S(\mathbf{t}, \mathcal{D})$\Comment{Generate solution with demonstrations}
                \If{$C(\mathbf{t}, \mathbf{s})$}
                    \State $\mathcal{T^*} \gets \mathcal{T^*} \cup (\mathbf{t}, \mathbf{s})$\Comment{Add to annotated tasks}
                    \State $\mathcal{U} \gets \mathcal{U} - \{\mathbf{t}\}$\Comment{Remove from unannotated tasks}
                \EndIf
            \EndFor
            \If{$\mathcal{U} = \emptyset$}
                \State \textbf{break}\Comment{All tasks annotated}
            \EndIf
        \EndFor
        \State \textbf{return} $\mathcal{T^*}$\Comment{Annotated tasks}
    \end{algorithmic}
\end{algorithm}

Given the success of demonstration selection for ICL for traditional NLP tasks, in this work, we explore how to effectively and efficiently leverage it for agentic tasks.
However, this requires a pool of tasks annotated with agent-style solutions\footnote{In real-world settings, these could also be obtained from an existing system.} (as described in \S~\ref{sec:agents}) that can serve as demonstrations. While some agentic benchmarks provide training sets of tasks, most do not provide task solutions in a form that can be used as demonstrations for the agent; rather, they only provide a final answer or a checker that can be used to verify solution correctness.

Since manually annotating tasks with solutions is intractable at scale, we propose a simple iterative algorithm (Algorithm \ref{alg:iterative}) to do this automatically.
Given (1) a pool of tasks $\mathcal{T}=\{\mathbf{t_i}\}_{i=1}^N$, (2) a solver $S$ that is used to generate solutions $\mathbf{s} \sim S(\mathbf{t}, \mathcal{D})$ given a task $\mathbf{t}$ and some demonstrations $\mathcal{D}$, and (3) a checker that verifies solution correctness, the algorithm returns tasks annotated with solutions $\mathcal{T^*}=\{\mathbf{t_i}, \mathbf{s_i}\}_{i=1}^N$.
Further, instead of naively retrying the solver, the algorithm also leverages currently annotated tasks as demonstrations. This not only improves efficiency in terms of the number of retries needed but also ensures that more instances are correctly annotated.

Finally, note that the algorithm, as described above, is agnostic to the kind of solver (and solutions), e.g., for the ReAct solver, the solutions would be trajectories, i.e. $\mathbf{s_i}=\tau_{i}$, while for the PnE solver, they would comprise a plan in the form of a sequence of subtasks and the corresponding trajectories, i.e. $\mathbf{s_i}=\{\mathbf{t_i^{j}}, \tau_{i}^{j}\}_{j=1}^{m_i}$. We will refer to the set of task trajectory annotations for ReAct as $\mathcal{D}_{\text{task}}^{*}$, the plan and subtask trajectory annotations for PnE as $\mathcal{D}_{\text{plan}}^{*}$ and $\mathcal{D}_{\text{subtask}}^{*}$, respectively.

\section{Demonstrations for Agents}
\label{sec:approach}

Having annotated a pool of tasks with solutions that can serve as demonstrations, we now discuss different demonstration granularities, along with how to select them, when to show them, and where to place them in the prompt.

\subsection{Task-level Trajectory Demonstrations}
\label{sec:traj}
Similar to traditional ICL, a natural way to show demonstrations is in the form of trajectories for similar tasks from the pool $\mathcal{D}_{\text{task}}^{*}$.
These task-trajectory pairs are selected from $\mathcal{D}_{\text{task}}^{*}$ before execution and are used in the prompt at every step. Specifically, they are placed in the general prompt $\textbf{p}$ before the description of the test task $\mathbf{t}$ and its execution trace $\mathbf{h_t}$.
To select these demonstrations, we experiment with both ranking-based and set-selection methods described in \S~\ref{sec:icl} using the task statement as the retrieval key.
Since, similarly to the tasks explored by \citet{gupta-etal-2023-coverage}, the optimal demonstrations for agentic tasks would demonstrate all necessary steps, we believe that set-selection might be more appropriate for agentic tasks as compared to ranking-based selection.

\subsection{Fine-grained Demonstrations}
As noted in \S~\ref{sec:agents}, the trajectories for complex agentic tasks can be very long. Thus, using them as demonstrations may be very expensive, if at all feasible with limited context lengths. Moreover, the trajectories for even the most similar tasks may have irrelevant steps while not being helpful for every step of the test task. Finally, LLMs have a recency bias \cite{liu-etal-2024-lost}, thus a \textit{smaller} demonstration, closer to the steps it is relevant for, may be more effective than an entire trajectory early in the prompt.
We explore two ways to achieve this: (1) using trajectories for similar subtasks with a PnE solver and (2) using snippets of the trajectories relevant to the current step.

\subsubsection{Subtask-level Trajectory Demonstrations}
\label{sec:subtask}
One way to use smaller demonstrations is to use the PnE solver instead of \S~\ref{sec:agents} that breaks the original task down to a sequence of subtasks $\mathbf{t^1}, \ldots, \mathbf{t^m}$ and executes each subtask separately using a React-style executor. The demonstrations for the PnE executor can be the trajectories of similar subtasks from the pool $\mathcal{D}_{\text{subtask}}^{*}$. This is similar to the trajectory demonstrations from the \S~\ref{sec:traj} in that they are selected prior to subtask execution and included in the general prompt $\mathbf{p}$ for every step of the subtask. However, the selection uses the subtask statement as the retrieval key, and demonstrations for different subtasks of the current task may be from different tasks. In this way, it allows for a more fine-grained demonstration to be shown for the limited scope of the subtask.

\subsubsection{Step-level Snippet Demonstrations}
\label{sec:snippet}
A major drawback of using a PnE solver is that it introduces a planning step prior to execution, which may yield inaccurate plans. Moreover, as they are placed early in the prompt, they still don't completely address the problem of recency bias.
Thus, we experiment with showing small snippets of trajectories that are relevant to the current step as demonstrations.
To identify these snippets, we use the thought produced at the current step as a retrieval key to find similar thoughts in trajectory annotations $\mathcal{D}_{\text{task}}^{*}$.
\begin{figure}[t]
    \centering
    \vspace{-1ex}
    \includegraphics[width=\linewidth]{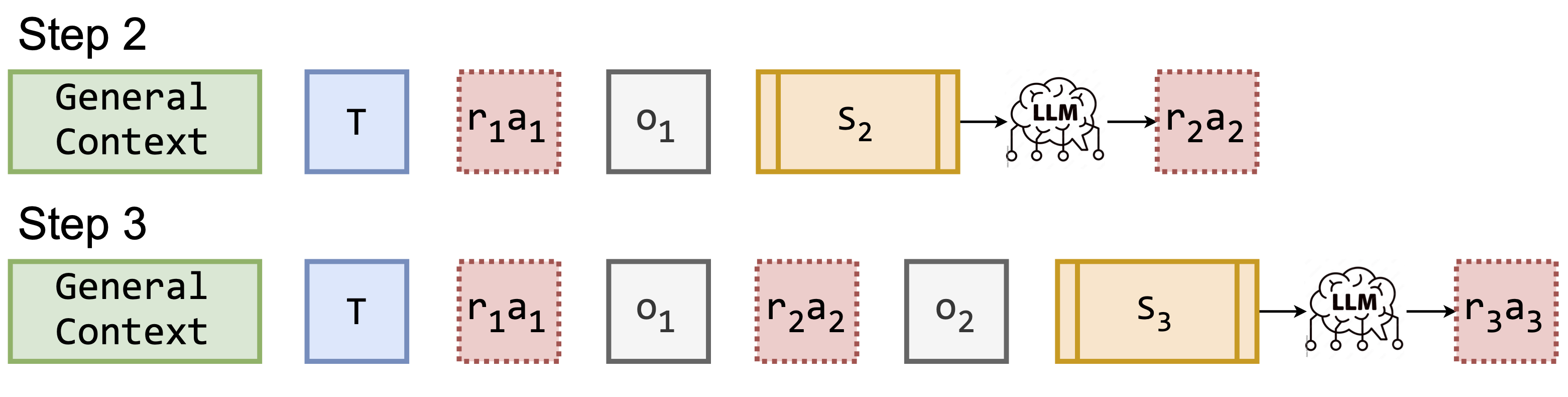}
    \caption{Snippet demonstrations are selected based on the thought at the current step (\textit{how}) and only used to predict the next thought-action (\textit{when}) by placing after the execution trace in the prompt (\textit{where}). E.g., the snippets $S_2$ are selected based on the thought $r_1$ and used to predict $r_2$ and $a_2$ and so on.}
    \label{fig:snippet_demo}
\end{figure}
A similar thought suggests a similar step, and we can use a snippet of the trajectory
comprising this step as a demonstration.
As shown in Fig. \ref{fig:snippet}, the selected snippets are appended to the prompt after the execution trace $\mathbf{h_t}$, which helps account for the recency bias. See Fig. \ref{fig:snippet_prompt} for an example of a prompt template for showing snippet demonstrations.
The snippets selected for the current step are only shown for predicting a single next step, as they may not be relevant for subsequent steps, for which we select new snippets anyway.
Finally, note that snippet demonstrations as described above do not require any additional annotations, as they are derived from the trajectory annotations.

\section{Experimental Setup}
\label{sec:exp_setup}

\subsection{AppWorld Benchmark}
\label{sec:appworld}

AppWorld\footnote{Our use of Appworld is permitted by its license (Apache License 2.0).} \cite{trivedi-etal-2024-appworld} is a benchmark designed to evaluate an autonomous agent's ability to carry out complex user tasks by interacting with 457 APIs associated with 9 simulated apps, viz. \textit{Gmail}, \textit{Venmo}, \textit{Spotify}, \textit{SimpleNote}, \textit{Splitwise}, \textit{Amazon}, \textit{Todoist}, \textit{Phone}, and \textit{File System}. It provides a stateful Python interpreter that the agent can use to interact with the various APIs and a \textit{Supervisor} app and an \textit{ApiDoc} app it can use to obtain the information about the user and the various Apps/APIs, respectively. Note that, although due to cost constraints, we only experiment with AppWorld, its code-based action space, varying observation sizes, and complex tasks make it the ideal testbed for our experiments.

The benchmark comprises a total of 244 task templates, or scenarios, each with three variants for a total of 722 tasks. The tasks are split into a train set (90 tasks), a dev set (57 tasks), and two test sets: test-normal (Test-N, 168 tasks), which evaluates in-distribution performance, and test-challenge (Test-C, 417 tasks), containing more complex tasks involving unseen apps.
Each task is associated with a suite of unit tests that check (1) whether only the requisite changes, and no extraneous changes, were made to the environment state, and (2) if required by the task, the final answer produced matches the ground truth. A task is considered solved only if all its unit tests pass.

\tightparagraph{Annotation.} To create our demonstration pool, we use the iterative annotation algorithm (Alg.~\S~\ref{alg:iterative}) to automatically annotate 146 tasks in the combined train and dev sets. As described in \S~\ref{sec:annotation}, we sped up the process by using already annotated instances as demonstrations for subsequent iterations.
Specifically, for ReAct, we used one task-trajectory pair as a demonstration, and for PnE, we used 4 task-plan pairs and 3 subtask-trajectory pairs for the planner and executor, respectively. For annotation, all the demonstrations were selected using Cosine Similarity based on \texttt{all-mpnet-base-v2} encoder. Of the 147 tasks, we annotated 141 tasks spanning 48 scenarios with ReAct solutions. With PnE, we were able to annotate 134 tasks spanning 46 scenarios. The PnE solutions had an average of ~6.2 subtasks per task for a total of 833 subtasks.

\tightparagraph{Evaluation.} We evaluate on both the Test-N and Test-C sets. AppWorld recommends two metrics: (1) Task Goal Completion (TGC), which is the percentage of tasks solved, and (2) Scenario Goal Completion (SGC), which is the percentage of scenarios for which all three task variants passed. While TGC measures an agent's overall performance, SGC measures its robustness across variations of a task.
Additionally, to assess whether agents solve tasks reliably, rather than by chance, we also report the percentage of tasks for which multiple \textit{runs} succeeded, called Reliable Task Goal Completion (RTGC).
We also report the average Token Usage during execution of each task as a measure of efficiency and the average number of Steps taken to complete tasks as a measure of inference costs. Note that token usage would aggregate the input and output token counts across all steps.

\subsection{Methods}
\label{sec:methods}

As discussed in \S~\ref{sec:approach}, we explore the following three different types of demonstrations:

\tightparagraph{Task Trajectory Demonstrations.} We experiment with the following selection methods to select $k$ trajectories: (1) ranking-based selection using Cosine Similarity (\textsc{Cos[$k$]}) and BertScore-Recall (\textsc{BSR[$k$]}), and (2) set-selection using \textsc{Set-BSR[$k$]} \cite{gupta-etal-2023-coverage}. Following \citet{gupta-etal-2023-coverage}, for \textsc{Cos}, we use \texttt{all-mpnet-base-v2} as the encoder, while for \textsc{BSR} and \textsc{Set-BSR}, we use \texttt{deberta-base-mnli-v2}. For \textsc{Cos} and \textsc{BSR}, we report results for $k=1$ and $k=2$, while for \textsc{Set-BSR}, we only report results for $k=2$ as selection using it reduces to \textsc{BSR} for $k=1$.
As baselines, we experiment with using the agent zero-shot without any trajectory demonstrations (\textsc{ZeroShot[0]}), and using a single fixed manually written trajectory from \citet{trivedi-etal-2024-appworld} as demonstration for every test input (\textsc{Fixed[1]}).

\tightparagraph{Subtask Trajectory Demonstrations.} We use \textsc{BSR} to select subtasks whose trajectories to include in the executor's prompt. Additionally, we use four task-plan pairs selected using \textsc{BSR} as demonstrations for the planner.

\tightparagraph{Snippet demonstrations.} We use \textsc{BSR} for selection of up to\footnote{To prevent spurious matches, we filter out thoughts that score less than 0.85.} $k=2$ annotated thoughts based on the thought at every step of the current task. For each selected thought, we create the snippets using the step (thought-action-observation triple) corresponding to the selected thought and a subsequent step if there is one.

\begin{resultstablesinglecol}[ht]
\begin{tabular}{lrrrrrr}
\toprule
Method & TGC $\uparrow$ & RTGC $\uparrow$ & SGC $\uparrow$ & Steps $\downarrow$ \\
\midrule
\textsc{ZeroShot[0]} & 35.1 & 22.0 & 15.2 & 21.9 \\
\midrule
\textsc{Fixed[1]} & 50.6 & 40.5 & 30.4 & 14.6 \\
\textsc{Random[1]} & 58.9 & 43.5 & 37.5 & 14.8 \\
\textsc{Cos[1]} & 64.0 & 57.1 & 44.6 & 13.4 \\
\textsc{BSR[1]} & 61.0 & 52.4 & 44.6 & 13.5 \\
\midrule
\textsc{Random[2]} & 58.9 & 45.8 & 33.9 & 13.8 \\
\textsc{Cos[2]} & 63.7 & 57.1 & 44.6 & 12.6 \\
\textsc{BSR[2]} & 64.9 & 59.5 & 50.0 & 12.2 \\
\textsc{SetBSR[2]} & \textbf{65.8} & \textbf{61.3} & \textbf{53.6} & \textbf{11.5} \\
\bottomrule
\end{tabular}
\caption{
Impact of different numbers and selection of trajectory demonstrations on a GPT-4o ReAct agent on the Test-N.
Even a single manually written trajectory significantly improves performance. Further, gains are obtained using actual agent trajectories, by selecting the most relevant trajectories as demonstrations, and by using set-selection when using multiple trajectories.
}
\label{tab:traj}
\end{resultstablesinglecol}

\subsection{Agent Implementation Details}
\label{sec:details}

We use OpenAI's GPT-4o (\texttt{gpt-4o-2024-08-06}) as the primary LLM both for annotating our demonstration pool as well as for our ICL experiments. Detailed hyperparameters and prompt templates for the ReAct solver, PnE planner, and executor are provided in the App. \ref{app:details}. To see if the annotation obtained from a larger LLM can benefit smaller LLMs, we also experiment with the smaller GPT-4o-mini (\texttt{gpt-4o-mini-2024-07-18}).

\begin{figure}
\centering
\includegraphics[width=0.9\linewidth]{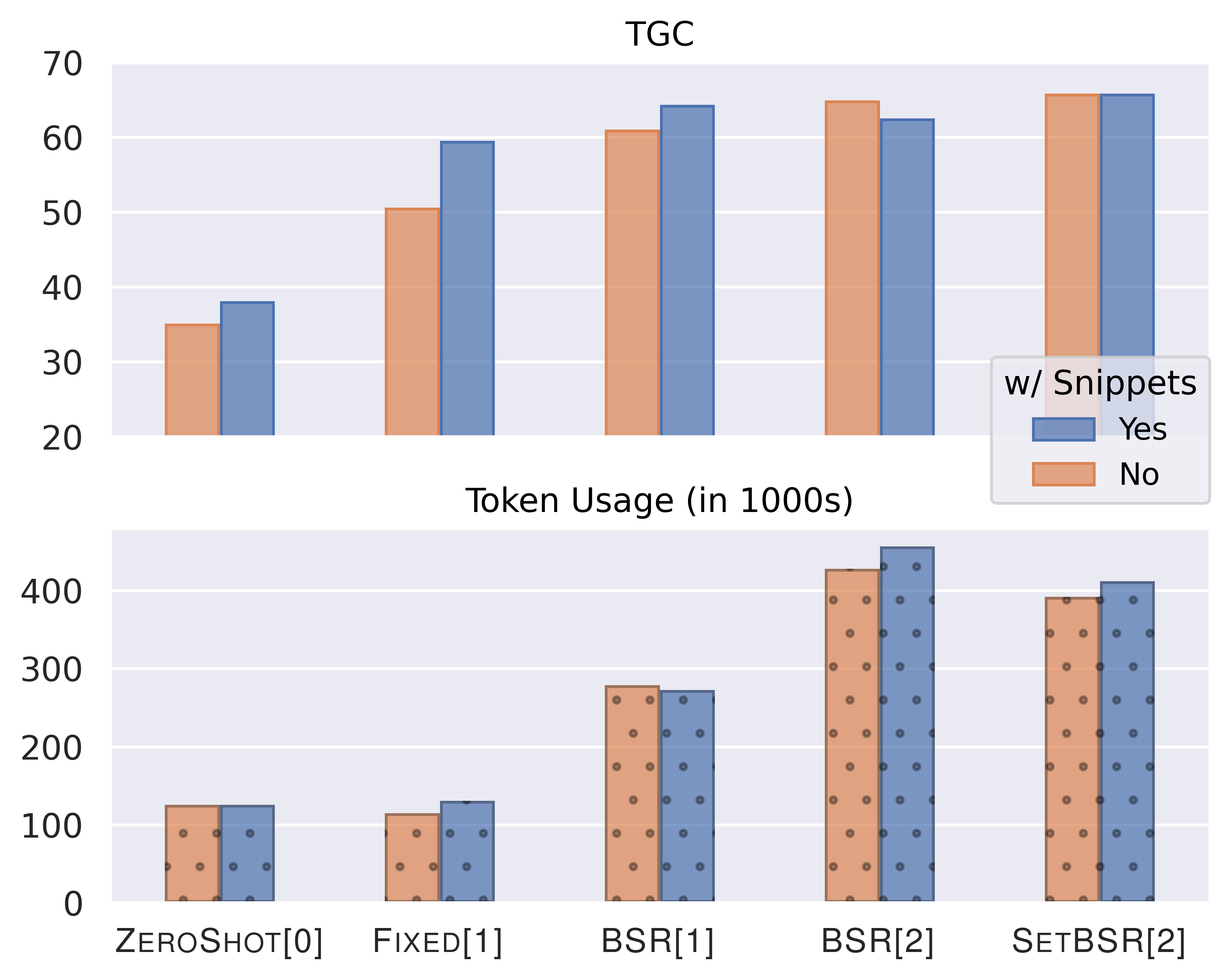}
\caption{Effect of trajectory and snippet demonstrations (selected using \textsc{BSR})
on the performance in terms of TGC (Top) and inference cost in terms of token usage per task (Bottom) of GPT-4o ReAct agents on Test-N. While trajectory demonstrations are most effective at improving performance, they do so at a high cost. Snippet demonstrations are also generally effective but have a very minimal overhead.}
\label{fig:snippet}
\end{figure}

\section{Results}
\label{sec:results}

\tightparagraph{Trajectory demonstrations boost agent performance.} Table \ref{tab:traj} shows the results on Test-N for varying numbers and selection of trajectory demonstrations.
First, it is clear from the TGC numbers (task completion) that even a single manually written trajectory demonstration (\textsc{Fixed}) can greatly improve agent performance compared to using the agent \textsc{ZeroShot}.
Moreover, the LLM agent's own trajectory annotations are more effective as demonstrations than simplified manually written ones, even if we select them randomly (\textsc{Random} v/s \textsc{Fixed}). Selecting a relevant trajectory, using any method \textsc{Cos} or \textsc{BSR}, remains the most effective. Finally, when using more than one trajectory, set-selection (\textsc{SetBSR}) is more effective than independent ranking-based selection.
Overall, \textsc{SetBSR[2]} beats \textsc{ZeroShot} and \textsc{Fixed} by 30.7 and 15.2 absolute points in TGC, respectively.

\tightparagraph{Trajectory demonstrations also make the agent more reliable, robust, and efficient.} Trajectory demonstrations have an even greater impact on RTGC and SGC (e.g.~\textsc{SetBSR[2]} improves on \textsc{Fixed}'s RTGC and SGC by 20.8 and 23.2 absolute points, respectively). This suggests that using relevant demonstrations is especially effective at making the agent more reliable (across multiple runs of the same task) and robust (across multiple variants of the task). Further, \textsc{SetBSR[2]} also takes 21\% fewer steps than \textsc{Fixed} and 47\% fewer steps than \textsc{ZeroShot}, implying greater efficiency and solution speed.

\tightparagraph{Snippet demonstrations generally improve performance with minimal overhead.} Fig.~\ref{fig:snippet} shows the results for the ReAct solver with and without snippet demonstrations for varying selections of trajectory demonstrations. Despite all their benefits, trajectories are very costly to use as demonstrations, and using two trajectories instead of one increases the average cost per task by 40\%. On the other hand, snippet demonstrations have very minimal overhead while generally improving performance. However, since they are not as effective as trajectories, the optimal approach is to use as many trajectories as possible and then sprinkle a few snippet demonstrations.

\begin{figure}[t]
\centering
\includegraphics[width=0.6\linewidth]{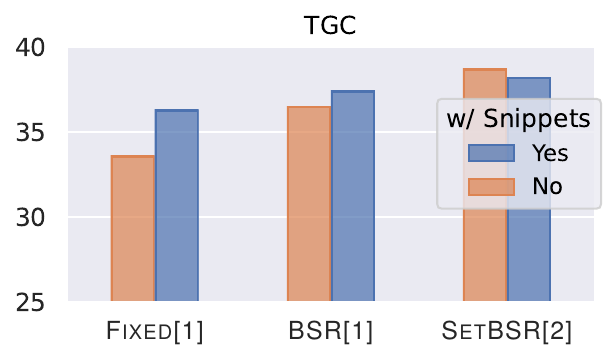}
\caption{Effect of trajectory and snippet demonstrations (selected using \textsc{BSR}) on TGC of GPT-4o ReAct agent on Test-C.}
\label{fig:tc}
\end{figure}

\tightparagraph{Demonstrations help even on out-of-domain tasks.}
As shown in Fig.~\ref{fig:tc}, demonstrations improve performance even on Test-C, which has more complex tasks involving unseen apps. As expected, the performance is lower than Test-N. However, although none of the annotations demonstrate the use of the unseen apps, we see that both trajectory and snippet demonstrations improve performance.

\begin{figure}[t]
\centering
\includegraphics[width=0.7\linewidth]{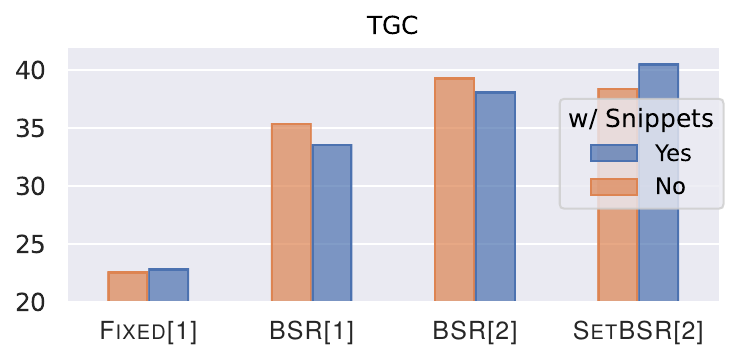}
\caption{Effect of trajectory and snippet demonstrations (selected using \textsc{BSR}) on TGC of GPT-4o-mini ReAct agent on Test-N.}
\label{fig:mini}
\end{figure}

\tightparagraph{Larger LLMs' annotations can also improve smaller LLM agents.}
As shown in Fig.~\ref{fig:mini}, GPT-4o's annotations also work well as demonstrations for the smaller GPT-4o-mini. As before, it's clear that both trajectory and snippet demonstrations are effective at improving performance and efficiency. Using \textsc{SetBSR[2]} trajectories with snippet demonstrations improves TGC rate by 14.3 absolute points and reduces the number of steps by 40\% compared to using just the \textsc{Fixed} trajectory demonstration.

\begin{figure}[t]
\centering
\includegraphics[width=0.9\linewidth]{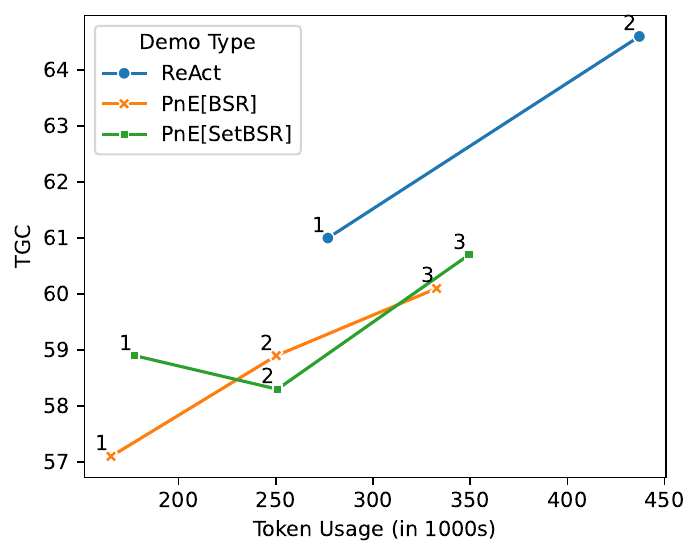}
\caption{Comparison of GPT-4o PnE (with \textsc{BSR} and \textsc{SetBSR} planner demos) and ReAct solvers with varying number of \textsc{BSR} trajectory demonstrations on Test-N. As trajectories for subtasks are much shorter than for entire tasks, more of the former can be used with a PnE executor than the latter with ReAct solver. However, PnE still underperforms ReAct likely because it attempts to decompose the task prior to any execution.}
\label{fig:pne}
\end{figure}

\tightparagraph{Subtask trajectory demonstrations improve PnE solver, but it still underperforms ReAct.}
Fig.~\ref{fig:pne} compares the PnE solver with \textsc{BSR} and \textsc{SetBSR}-selected planner demonstrations and the ReAct solver for a varying number of trajectory demonstrations. It is clear that subtask trajectories have much less overhead than task trajectories. Nevertheless, despite adding more trajectory demonstrations for the executor, the PnE solver cannot match the ReAct solver. This is likely because PnE plans before any execution, which may lead to inaccurate plans.

\tightparagraph{With demonstration selection prompted agents can be competitive with trained agents.} Table \ref{tab:trained} compares the GPT-4o ReAct agent with a variety of supervised finetuning (SFT), direct preference optimization (DPO), reinforcement learning (RL) approaches to train \texttt{Qwen-2.5-32B}-based agents from \citet{loop}. We provide a brief description of each approach in App. \ref{app:trained} and refer the reader to \citet{loop} for more details. It is clear that with selected trajectory and snippet demonstrations, GPT-4o ReAct agent can outperform all but the best-trained agents.

\begin{resultstablesinglecol}[tbp]
\begin{tabular}{@{}clcccc@{}}
\toprule
\multicolumn{2}{c}{\multirow{2}{*}{Approach}} & \multicolumn{2}{c}{Test-N} & \multicolumn{2}{c}{Test-C} \\ \cmidrule(lr){3-4}\cmidrule(lr){5-6}
    & & TGC & SGC & TGC & SGC \\ \midrule
\multirow{3}{*}{SFT}
    & SFT-GT & 6.2 & 1.8 & 0.8 & 0.1 \\
    & RFT & 47.9 & 26.4 & 26.4 & 11.4 \\
    & EI & 58.3 & 36.8 & 32.8 & 17.6\\ \midrule
\multirow{2}{*}{DPO}
    & DPO-MCTS & 57.0 & 31.8 & 31.8 & 13.7\\
    & DMPO & 59.0 & 36.6 & 36.3 & 13.7\\ \midrule
\multirow{6}{*}{RL}
    & PPO (learned critic, token) & 50.8 & 28.9 & 26.4	& 10.5 \\
    & RLOO (traj) & 57.2 & 35.7 & 36.7	& 17.4 \\
    & GRPO (token) & 58.0 & 36.8 & 39.5	& 22.4 \\
    & LOOP (bandit) & 53.3 & 33.6 & 27.7 & 13.0 \\
    & LOOP (turn) & 64.1 & 43.5 & 40.8	& 26.5 \\
    & LOOP (token) & \textbf{71.3} & \textbf{53.6} & \textbf{45.7}	& \textbf{26.6} \\ \midrule
\multirow{3}{*}{NFT}
    & Traj[Fixed] & 50.6 & 30.4 & 33.6 & 18.0 \\
    & Traj[SetBSR] & 65.8 & \textbf{53.6} & 38.7 & 24.8 \\
    & Traj[SetBSR]+Snippet & 65.8 & \textbf{53.6} & 38.2 & 23.4 \\ \bottomrule
\end{tabular}
\caption{
    With selected trajectory and snippet demonstrations, a prompted GPT-4o agent performs competitively with \texttt{Qwen-2.5-32B}-based agents trained using a variety of approaches spanning supervised finetuning (SFT), direct preference optimization (DPO), and reinforcement learning (RL). The results for trained agents are taken from \citet{loop}. We refer the reader to App. \ref{app:trained} for a brief description of each approach and to \citet{loop} for more details.
    }
\label{tab:trained}
\end{resultstablesinglecol}

\section{Conclusion}

This work studied different design decisions relating to ICL with demonstration selection for LLM agents. We proposed a novel iterative annotation algorithm to automatically annotate training tasks with solutions for use as demonstrations. Using these annotations, we showed that trajectory demonstrations can effectively improve performance, reliability, robustness, and efficiency of LLM agents. Further, since trajectory demonstrations can have a large overhead in terms of inference costs, we also showed that small snippets of trajectories can be used as demonstrations at every step to boost performance with a minimal overhead.
Overall, our results suggest that the optimal ICL approach is to use as many trajectory demonstrations as possible and then sprinkle a few snippets, and that this can yield prompted LLM agents that are competitive with state-of-the-art trained agents. 

\section*{Acknowledgements}
This work was funded in part by the DARPA ANSR program under award FA8750-23-2-0004, in part by NSF CAREER award \#IIS-2046873, and in part by NSF CCRI award \#CNS-1925741. The views expressed are those of the authors and do not reflect the policy of the funding agencies.

\section*{Limitations}
In this work, we were primarily concerned with studying the right form and placement of demonstrations, but used general-purpose encoders for their selection. Further, retrieval-based approaches select demonstrations based on a retrieval key (e.g., task statements) and hence cannot take advantage of additional solution information that may only be available for the demonstration candidates but not for the test task. Future work can explore thus explore demonstration selection approaches more suitable for agentic tasks.

\bibliography{bibliography/anthology-part1,bibliography/anthology-part2,bibliography/custom-rebiber}

\begin{thebibliography}{40}
\providecommand{\natexlab}[1]{#1}

\bibitem[{Agrawal et~al.(2023)Agrawal, Zhou, Lewis, Zettlemoyer, and Ghazvininejad}]{agrawal2022incontext}
Sweta Agrawal, Chunting Zhou, Mike Lewis, Luke Zettlemoyer, and Marjan Ghazvininejad. 2023.
\newblock \href {https://doi.org/10.18653/v1/2023.findings-acl.564} {In-context examples selection for machine translation}.
\newblock In \emph{Findings of the Association for Computational Linguistics: ACL 2023}, pages 8857--8873, Toronto, Canada. Association for Computational Linguistics.

\bibitem[{Ahmadian et~al.(2024)Ahmadian, Cremer, Gallé, Fadaee, Kreutzer, Pietquin, Üstün, and Hooker}]{rloo}
Arash Ahmadian, Chris Cremer, Matthias Gallé, Marzieh Fadaee, Julia Kreutzer, Olivier Pietquin, Ahmet Üstün, and Sara Hooker. 2024.
\newblock \href {https://arxiv.org/abs/2402.14740} {Back to basics: Revisiting reinforce style optimization for learning from human feedback in llms}.

\bibitem[{Anthony et~al.(2017)Anthony, Tian, and Barber}]{ei}
Thomas Anthony, Zheng Tian, and David Barber. 2017.
\newblock \href {https://proceedings.neurips.cc/paper/2017/hash/d8e1344e27a5b08cdfd5d027d9b8d6de-Abstract.html} {Thinking fast and slow with deep learning and tree search}.
\newblock In \emph{Advances in Neural Information Processing Systems 30: Annual Conference on Neural Information Processing Systems 2017, December 4-9, 2017, Long Beach, CA, {USA}}, pages 5360--5370.

\bibitem[{Askari et~al.(2025)Askari, Gupta, Tong, Wang, Chhabra, and Chen}]{askari2025unraveling}
Hadi Askari, Shivanshu Gupta, Terry Tong, Fei Wang, Anshuman Chhabra, and Muhao Chen. 2025.
\newblock \href {https://arxiv.org/abs/2501.01473} {Unraveling indirect in-context learning using influence functions}.

\bibitem[{Bogin et~al.(2024)Bogin, Yang, Gupta, Richardson, Bransom, Clark, Sabharwal, and Khot}]{bogin2024super}
Ben Bogin, Kejuan Yang, Shashank Gupta, Kyle Richardson, Erin Bransom, Peter Clark, Ashish Sabharwal, and Tushar Khot. 2024.
\newblock \href {https://arxiv.org/abs/2409.07440} {Super: Evaluating agents on setting up and executing tasks from research repositories}.

\bibitem[{Brown et~al.(2020)Brown, Mann, Ryder, Subbiah, Kaplan, Dhariwal, Neelakantan, Shyam, Sastry, Askell, Agarwal, Herbert{-}Voss, Krueger, Henighan, Child, Ramesh, Ziegler, Wu, Winter, Hesse, Chen, Sigler, Litwin, Gray, Chess, Clark, Berner, McCandlish, Radford, Sutskever, and Amodei}]{brown-etal-2020-fewshot}
Tom~B. Brown, Benjamin Mann, Nick Ryder, Melanie Subbiah, Jared Kaplan, Prafulla Dhariwal, Arvind Neelakantan, Pranav Shyam, Girish Sastry, Amanda Askell, Sandhini Agarwal, Ariel Herbert{-}Voss, Gretchen Krueger, Tom Henighan, Rewon Child, Aditya Ramesh, Daniel~M. Ziegler, Jeffrey Wu, Clemens Winter, and 12 others. 2020.
\newblock \href {https://proceedings.neurips.cc/paper/2020/hash/1457c0d6bfcb4967418bfb8ac142f64a-Abstract.html} {Language models are few-shot learners}.
\newblock In \emph{Advances in Neural Information Processing Systems 33: Annual Conference on Neural Information Processing Systems 2020, NeurIPS 2020, December 6-12, 2020, virtual}.

\bibitem[{Chen et~al.(2023)Chen, Shu, Shareghi, Collier, Narasimhan, and Yao}]{chen2023fireact}
Baian Chen, Chang Shu, Ehsan Shareghi, Nigel Collier, Karthik Narasimhan, and Shunyu Yao. 2023.
\newblock \href {https://arxiv.org/abs/2310.05915} {Fireact: Toward language agent fine-tuning}.

\bibitem[{Chen et~al.(2025)Chen, Cusumano-Towner, Huval, Petrenko, Hamburger, Koltun, and Krähenbühl}]{loop}
Kevin Chen, Marco Cusumano-Towner, Brody Huval, Aleksei Petrenko, Jackson Hamburger, Vladlen Koltun, and Philipp Krähenbühl. 2025.
\newblock \href {https://arxiv.org/abs/2502.01600} {Reinforcement learning for long-horizon interactive llm agents}.

\bibitem[{Deng et~al.(2023)Deng, Gu, Zheng, Chen, Stevens, Wang, Sun, and Su}]{deng2023mind2web}
Xiang Deng, Yu~Gu, Boyuan Zheng, Shijie Chen, Samual Stevens, Boshi Wang, Huan Sun, and Yu~Su. 2023.
\newblock \href {http://papers.nips.cc/paper\_files/paper/2023/hash/5950bf290a1570ea401bf98882128160-Abstract-Datasets\_and\_Benchmarks.html} {Mind2web: Towards a generalist agent for the web}.
\newblock In \emph{Advances in Neural Information Processing Systems 36: Annual Conference on Neural Information Processing Systems 2023, NeurIPS 2023, New Orleans, LA, USA, December 10 - 16, 2023}.

\bibitem[{Drouin et~al.(2024)Drouin, Gasse, Caccia, Laradji, Verme, Marty, V{\'{a}}zquez, Chapados, and Lacoste}]{drouin2024workarena}
Alexandre Drouin, Maxime Gasse, Massimo Caccia, Issam~H. Laradji, Manuel~Del Verme, Tom Marty, David V{\'{a}}zquez, Nicolas Chapados, and Alexandre Lacoste. 2024.
\newblock \href {https://openreview.net/forum?id=BRfqYrikdo} {Workarena: How capable are web agents at solving common knowledge work tasks?}
\newblock In \emph{Forty-first International Conference on Machine Learning, {ICML} 2024, Vienna, Austria, July 21-27, 2024}. OpenReview.net.

\bibitem[{Gupta et~al.(2023)Gupta, Gardner, and Singh}]{gupta-etal-2023-coverage}
Shivanshu Gupta, Matt Gardner, and Sameer Singh. 2023.
\newblock \href {https://doi.org/10.18653/v1/2023.findings-emnlp.930} {Coverage-based example selection for in-context learning}.
\newblock In \emph{Findings of the Association for Computational Linguistics: EMNLP 2023}, pages 13924--13950, Singapore. Association for Computational Linguistics.

\bibitem[{Gupta et~al.(2024)Gupta, Rosenbaum, and Elenberg}]{gistscore}
Shivanshu Gupta, Clemens Rosenbaum, and Ethan~R. Elenberg. 2024.
\newblock \href {https://openreview.net/forum?id=WCVC5wGZyz} {Gistscore: Learning better representations for in-context example selection with gist bottlenecks}.
\newblock In \emph{Forty-first International Conference on Machine Learning, {ICML} 2024, Vienna, Austria, July 21-27, 2024}. OpenReview.net.

\bibitem[{Kim et~al.(2023)Kim, Baldi, and McAleer}]{kim2023language}
Geunwoo Kim, Pierre Baldi, and Stephen McAleer. 2023.
\newblock \href {http://papers.nips.cc/paper\_files/paper/2023/hash/7cc1005ec73cfbaac9fa21192b622507-Abstract-Conference.html} {Language models can solve computer tasks}.
\newblock In \emph{Advances in Neural Information Processing Systems 36: Annual Conference on Neural Information Processing Systems 2023, NeurIPS 2023, New Orleans, LA, USA, December 10 - 16, 2023}.

\bibitem[{Levy et~al.(2023)Levy, Bogin, and Berant}]{levy-etal-2023-diverse}
Itay Levy, Ben Bogin, and Jonathan Berant. 2023.
\newblock \href {https://doi.org/10.18653/v1/2023.acl-long.78} {Diverse demonstrations improve in-context compositional generalization}.
\newblock In \emph{Proceedings of the 61st Annual Meeting of the Association for Computational Linguistics (Volume 1: Long Papers)}, pages 1401--1422, Toronto, Canada. Association for Computational Linguistics.

\bibitem[{Liu et~al.(2022)Liu, Shen, Zhang, Dolan, Carin, and Chen}]{liu-etal-2022-makes}
Jiachang Liu, Dinghan Shen, Yizhe Zhang, Bill Dolan, Lawrence Carin, and Weizhu Chen. 2022.
\newblock \href {https://doi.org/10.18653/v1/2022.deelio-1.10} {What makes good in-context examples for {GPT}-3?}
\newblock In \emph{Proceedings of Deep Learning Inside Out (DeeLIO 2022): The 3rd Workshop on Knowledge Extraction and Integration for Deep Learning Architectures}, pages 100--114, Dublin, Ireland and Online. Association for Computational Linguistics.

\bibitem[{Liu et~al.(2024)Liu, Lin, Hewitt, Paranjape, Bevilacqua, Petroni, and Liang}]{liu-etal-2024-lost}
Nelson~F. Liu, Kevin Lin, John Hewitt, Ashwin Paranjape, Michele Bevilacqua, Fabio Petroni, and Percy Liang. 2024.
\newblock \href {https://doi.org/10.1162/tacl_a_00638} {Lost in the middle: How language models use long contexts}.
\newblock \emph{Transactions of the Association for Computational Linguistics}, 12:157--173.

\bibitem[{Mitra et~al.(2024)Mitra, Corro, Zheng, Mahajan, Rouhana, Codas, Lu, ge~Chen, Vrousgos, Rosset, Silva, Khanpour, Lara, and Awadallah}]{mitra2024agentinstruct}
Arindam Mitra, Luciano~Del Corro, Guoqing Zheng, Shweti Mahajan, Dany Rouhana, Andres Codas, Yadong Lu, Wei ge~Chen, Olga Vrousgos, Corby Rosset, Fillipe Silva, Hamed Khanpour, Yash Lara, and Ahmed Awadallah. 2024.
\newblock \href {https://arxiv.org/abs/2407.03502} {Agentinstruct: Toward generative teaching with agentic flows}.

\bibitem[{Nakano et~al.(2021)Nakano, Hilton, Balaji, Wu, Ouyang, Kim, Hesse, Jain, Kosaraju, Saunders, Jiang, Cobbe, Eloundou, Krueger, Button, Knight, Chess, and Schulman}]{nakano2022webgpt}
Reiichiro Nakano, Jacob Hilton, Suchir Balaji, Jeff Wu, Long Ouyang, Christina Kim, Christopher Hesse, Shantanu Jain, Vineet Kosaraju, William Saunders, Xu~Jiang, Karl Cobbe, Tyna Eloundou, Gretchen Krueger, Kevin Button, Matthew Knight, Benjamin Chess, and John Schulman. 2021.
\newblock \href {https://arxiv.org/abs/2112.09332} {Webgpt: Browser-assisted question-answering with human feedback}.

\bibitem[{Putta et~al.(2024)Putta, Mills, Garg, Motwani, Finn, Garg, and Rafailov}]{dpo}
Pranav Putta, Edmund Mills, Naman Garg, Sumeet Motwani, Chelsea Finn, Divyansh Garg, and Rafael Rafailov. 2024.
\newblock \href {https://arxiv.org/abs/2408.07199} {Agent q: Advanced reasoning and learning for autonomous ai agents}.

\bibitem[{Qin et~al.(2024)Qin, Liang, Ye, Zhu, Yan, Lu, Lin, Cong, Tang, Qian, Zhao, Hong, Tian, Xie, Zhou, Gerstein, Li, Liu, and Sun}]{qin2023toolllm}
Yujia Qin, Shihao Liang, Yining Ye, Kunlun Zhu, Lan Yan, Yaxi Lu, Yankai Lin, Xin Cong, Xiangru Tang, Bill Qian, Sihan Zhao, Lauren Hong, Runchu Tian, Ruobing Xie, Jie Zhou, Mark Gerstein, Dahai Li, Zhiyuan Liu, and Maosong Sun. 2024.
\newblock \href {https://openreview.net/forum?id=dHng2O0Jjr} {Toolllm: Facilitating large language models to master 16000+ real-world apis}.
\newblock In \emph{The Twelfth International Conference on Learning Representations, {ICLR} 2024, Vienna, Austria, May 7-11, 2024}. OpenReview.net.

\bibitem[{Rubin et~al.(2022)Rubin, Herzig, and Berant}]{rubin-etal-2022-learning}
Ohad Rubin, Jonathan Herzig, and Jonathan Berant. 2022.
\newblock \href {https://doi.org/10.18653/v1/2022.naacl-main.191} {Learning to retrieve prompts for in-context learning}.
\newblock In \emph{Proceedings of the 2022 Conference of the North American Chapter of the Association for Computational Linguistics: Human Language Technologies}, pages 2655--2671, Seattle, United States. Association for Computational Linguistics.

\bibitem[{Schulman et~al.(2017)Schulman, Wolski, Dhariwal, Radford, and Klimov}]{ppo}
John Schulman, Filip Wolski, Prafulla Dhariwal, Alec Radford, and Oleg Klimov. 2017.
\newblock \href {https://arxiv.org/abs/1707.06347} {Proximal policy optimization algorithms}.

\bibitem[{Shao et~al.(2024)Shao, Wang, Zhu, Xu, Song, Bi, Zhang, Zhang, Li, Wu, and Guo}]{grpo}
Zhihong Shao, Peiyi Wang, Qihao Zhu, Runxin Xu, Junxiao Song, Xiao Bi, Haowei Zhang, Mingchuan Zhang, Y.~K. Li, Y.~Wu, and Daya Guo. 2024.
\newblock \href {https://arxiv.org/abs/2402.03300} {Deepseekmath: Pushing the limits of mathematical reasoning in open language models}.

\bibitem[{Shinn et~al.(2023)Shinn, Cassano, Gopinath, Narasimhan, and Yao}]{shinn2023reflexion}
Noah Shinn, Federico Cassano, Ashwin Gopinath, Karthik Narasimhan, and Shunyu Yao. 2023.
\newblock \href {http://papers.nips.cc/paper\_files/paper/2023/hash/1b44b878bb782e6954cd888628510e90-Abstract-Conference.html} {Reflexion: language agents with verbal reinforcement learning}.
\newblock In \emph{Advances in Neural Information Processing Systems 36: Annual Conference on Neural Information Processing Systems 2023, NeurIPS 2023, New Orleans, LA, USA, December 10 - 16, 2023}.

\bibitem[{Shridhar et~al.(2021)Shridhar, Yuan, C{\^{o}}t{\'{e}}, Bisk, Trischler, and Hausknecht}]{shridhar2021alfworld}
Mohit Shridhar, Xingdi Yuan, Marc{-}Alexandre C{\^{o}}t{\'{e}}, Yonatan Bisk, Adam Trischler, and Matthew~J. Hausknecht. 2021.
\newblock \href {https://openreview.net/forum?id=0IOX0YcCdTn} {Alfworld: Aligning text and embodied environments for interactive learning}.
\newblock In \emph{9th International Conference on Learning Representations, {ICLR} 2021, Virtual Event, Austria, May 3-7, 2021}. OpenReview.net.

\bibitem[{Su et~al.(2023)Su, Kasai, Wu, Shi, Wang, Xin, Zhang, Ostendorf, Zettlemoyer, Smith, and Yu}]{su2022selective}
Hongjin Su, Jungo Kasai, Chen~Henry Wu, Weijia Shi, Tianlu Wang, Jiayi Xin, Rui Zhang, Mari Ostendorf, Luke Zettlemoyer, Noah~A. Smith, and Tao Yu. 2023.
\newblock \href {https://openreview.net/pdf?id=qY1hlv7gwg} {Selective annotation makes language models better few-shot learners}.
\newblock In \emph{The Eleventh International Conference on Learning Representations, {ICLR} 2023, Kigali, Rwanda, May 1-5, 2023}. OpenReview.net.

\bibitem[{Sun et~al.(2024)Sun, Liu, Wang, Iter, Zhu, and Iyyer}]{sun-etal-2024-pearl}
Simeng Sun, Yang Liu, Shuohang Wang, Dan Iter, Chenguang Zhu, and Mohit Iyyer. 2024.
\newblock \href {https://aclanthology.org/2024.eacl-long.29/} {{PEARL}: Prompting large language models to plan and execute actions over long documents}.
\newblock In \emph{Proceedings of the 18th Conference of the European Chapter of the Association for Computational Linguistics (Volume 1: Long Papers)}, pages 469--486, St. Julian{'}s, Malta. Association for Computational Linguistics.

\bibitem[{Trivedi et~al.(2024)Trivedi, Khot, Hartmann, Manku, Dong, Li, Gupta, Sabharwal, and Balasubramanian}]{trivedi-etal-2024-appworld}
Harsh Trivedi, Tushar Khot, Mareike Hartmann, Ruskin Manku, Vinty Dong, Edward Li, Shashank Gupta, Ashish Sabharwal, and Niranjan Balasubramanian. 2024.
\newblock \href {https://doi.org/10.18653/v1/2024.acl-long.850} {{A}pp{W}orld: A controllable world of apps and people for benchmarking interactive coding agents}.
\newblock In \emph{Proceedings of the 62nd Annual Meeting of the Association for Computational Linguistics (Volume 1: Long Papers)}, pages 16022--16076, Bangkok, Thailand. Association for Computational Linguistics.

\bibitem[{Wang et~al.(2023)Wang, Cai, Chen, Liu, Ma, and Liang}]{wang2024deps}
Zihao Wang, Shaofei Cai, Guanzhou Chen, Anji Liu, Xiaojian Ma, and Yitao Liang. 2023.
\newblock \href {https://arxiv.org/abs/2302.01560} {Describe, explain, plan and select: Interactive planning with large language models enables open-world multi-task agents}.

\bibitem[{Yang et~al.(2023)Yang, Prabhakar, Narasimhan, and Yao}]{yang2023intercode}
John Yang, Akshara Prabhakar, Karthik Narasimhan, and Shunyu Yao. 2023.
\newblock \href {http://papers.nips.cc/paper\_files/paper/2023/hash/4b175d846fb008d540d233c188379ff9-Abstract-Datasets\_and\_Benchmarks.html} {Intercode: Standardizing and benchmarking interactive coding with execution feedback}.
\newblock In \emph{Advances in Neural Information Processing Systems 36: Annual Conference on Neural Information Processing Systems 2023, NeurIPS 2023, New Orleans, LA, USA, December 10 - 16, 2023}.

\bibitem[{Yao et~al.(2022)Yao, Chen, Yang, and Narasimhan}]{yao2023webshop}
Shunyu Yao, Howard Chen, John Yang, and Karthik Narasimhan. 2022.
\newblock \href {http://papers.nips.cc/paper\_files/paper/2022/hash/82ad13ec01f9fe44c01cb91814fd7b8c-Abstract-Conference.html} {Webshop: Towards scalable real-world web interaction with grounded language agents}.
\newblock In \emph{Advances in Neural Information Processing Systems 35: Annual Conference on Neural Information Processing Systems 2022, NeurIPS 2022, New Orleans, LA, USA, November 28 - December 9, 2022}.

\bibitem[{Yao et~al.(2023)Yao, Zhao, Yu, Du, Shafran, Narasimhan, and Cao}]{react}
Shunyu Yao, Jeffrey Zhao, Dian Yu, Nan Du, Izhak Shafran, Karthik~R. Narasimhan, and Yuan Cao. 2023.
\newblock \href {https://openreview.net/pdf?id=WE\_vluYUL-X} {React: Synergizing reasoning and acting in language models}.
\newblock In \emph{The Eleventh International Conference on Learning Representations, {ICLR} 2023, Kigali, Rwanda, May 1-5, 2023}. OpenReview.net.

\bibitem[{Ye et~al.(2023{\natexlab{a}})Ye, Wu, Feng, Yu, and Kong}]{ye2023compositional}
Jiacheng Ye, Zhiyong Wu, Jiangtao Feng, Tao Yu, and Lingpeng Kong. 2023{\natexlab{a}}.
\newblock \href {https://proceedings.mlr.press/v202/ye23c.html} {Compositional exemplars for in-context learning}.
\newblock In \emph{International Conference on Machine Learning, {ICML} 2023, 23-29 July 2023, Honolulu, Hawaii, {USA}}, volume 202 of \emph{Proceedings of Machine Learning Research}, pages 39818--39833. {PMLR}.

\bibitem[{Ye et~al.(2023{\natexlab{b}})Ye, Iyer, Celikyilmaz, Stoyanov, Durrett, and Pasunuru}]{ye2022complementary}
Xi~Ye, Srinivasan Iyer, Asli Celikyilmaz, Veselin Stoyanov, Greg Durrett, and Ramakanth Pasunuru. 2023{\natexlab{b}}.
\newblock \href {https://doi.org/10.18653/v1/2023.findings-acl.273} {Complementary explanations for effective in-context learning}.
\newblock In \emph{Findings of the Association for Computational Linguistics: ACL 2023}, pages 4469--4484, Toronto, Canada. Association for Computational Linguistics.

\bibitem[{Yuan et~al.(2023)Yuan, Yuan, Li, Dong, Lu, Tan, Zhou, and Zhou}]{rft}
Zheng Yuan, Hongyi Yuan, Chengpeng Li, Guanting Dong, Keming Lu, Chuanqi Tan, Chang Zhou, and Jingren Zhou. 2023.
\newblock \href {https://arxiv.org/abs/2308.01825} {Scaling relationship on learning mathematical reasoning with large language models}.

\bibitem[{Zhang et~al.(2020)Zhang, Kishore, Wu, Weinberger, and Artzi}]{zhang2020bertscore}
Tianyi Zhang, Varsha Kishore, Felix Wu, Kilian~Q. Weinberger, and Yoav Artzi. 2020.
\newblock \href {https://openreview.net/forum?id=SkeHuCVFDr} {Bertscore: Evaluating text generation with {BERT}}.
\newblock In \emph{8th International Conference on Learning Representations, {ICLR} 2020, Addis Ababa, Ethiopia, April 26-30, 2020}. OpenReview.net.

\bibitem[{Zhao et~al.(2021)Zhao, Wallace, Feng, Klein, and Singh}]{zhao2021calibrate}
Zihao Zhao, Eric Wallace, Shi Feng, Dan Klein, and Sameer Singh. 2021.
\newblock \href {http://proceedings.mlr.press/v139/zhao21c.html} {Calibrate before use: Improving few-shot performance of language models}.
\newblock In \emph{Proceedings of the 38th International Conference on Machine Learning, {ICML} 2021, 18-24 July 2021, Virtual Event}, volume 139 of \emph{Proceedings of Machine Learning Research}, pages 12697--12706. {PMLR}.

\bibitem[{Zheng et~al.(2024)Zheng, Wang, Wang, and An}]{synapse}
Longtao Zheng, Rundong Wang, Xinrun Wang, and Bo~An. 2024.
\newblock \href {https://openreview.net/forum?id=Pc8AU1aF5e} {Synapse: Trajectory-as-exemplar prompting with memory for computer control}.
\newblock In \emph{The Twelfth International Conference on Learning Representations, {ICLR} 2024, Vienna, Austria, May 7-11, 2024}. OpenReview.net.

\bibitem[{Zhou et~al.(2024{\natexlab{a}})Zhou, Yang, Wen, Wen, Wang, Xi, Xu, Yu, and Zhang}]{trad}
Ruiwen Zhou, Yingxuan Yang, Muning Wen, Ying Wen, Wenhao Wang, Chunling Xi, Guoqiang Xu, Yong Yu, and Weinan Zhang. 2024{\natexlab{a}}.
\newblock \href {https://doi.org/10.1145/3626772.3657788} {{TRAD:} enhancing {LLM} agents with step-wise thought retrieval and aligned decision}.
\newblock In \emph{Proceedings of the 47th International {ACM} {SIGIR} Conference on Research and Development in Information Retrieval, {SIGIR} 2024, Washington DC, USA, July 14-18, 2024}, pages 3--13. {ACM}.

\bibitem[{Zhou et~al.(2024{\natexlab{b}})Zhou, Xu, Zhu, Zhou, Lo, Sridhar, Cheng, Ou, Bisk, Fried, Alon, and Neubig}]{zhou2024webarena}
Shuyan Zhou, Frank~F. Xu, Hao Zhu, Xuhui Zhou, Robert Lo, Abishek Sridhar, Xianyi Cheng, Tianyue Ou, Yonatan Bisk, Daniel Fried, Uri Alon, and Graham Neubig. 2024{\natexlab{b}}.
\newblock \href {https://openreview.net/forum?id=oKn9c6ytLx} {Webarena: {A} realistic web environment for building autonomous agents}.
\newblock In \emph{The Twelfth International Conference on Learning Representations, {ICLR} 2024, Vienna, Austria, May 7-11, 2024}. OpenReview.net.

\end{thebibliography}

\clearpage

\appendix
\section{Agent Details}
\label{app:details}

Table \ref{tab:hyperparams} shows various hyperparameters used for a ReAct solver and PnE executor when solving a task or subtask, respectively.

\subsection{Prompt Formats}

\tightparagraph{ReAct Solver} Fig.~\ref{fig:react_prompt} shows the format of the task context we use for the ReAct solver. It also shows the JSON format in which the agent is constrained to generate its reasoning and action. The hyperparameters used for the ReAct solver are given in Table \ref{tab:hyperparams}.

\tightparagraph{PnE Planner} The prompt template for the planner is given in Fig. \ref{fig:planner_prompt}. The planner is also constrained to generate the plan using a JSON format. For the planner, we use \texttt{temperature} 0.1 and \texttt{top\_p} 0.5 both during annotation and evaluation.

\tightparagraph{PnE Executor} The PnE executor is similar to the ReAct solver. The format of the task context used for the PnE executor is given in Fig. \ref{fig:executor_prompt1}. A The executor is also constrained to generate its reasoning and using the same JSON format as the ReAct solver. The hyperparameters used for the PnE executor are given in Table \ref{tab:hyperparams}.

\tightparagraph{Prompt Truncation} When the ReAct solver or PnE executor's prompt exceeds the corresponding context length limit, we truncated it by first hiding the older and longer observations and then hiding any remaining older observations.

\subsection{Demonstration Templates}
The templates used to show the task-trajectory, subtask-trajectory, and snippet demonstrations are given in Figs. \ref{fig:traj_prompt}, \ref{fig:subtraj_prompt}, and \ref{fig:snippet_prompt}, respectively.

\begin{resultstable}[tbp]
    \begin{tabular}{lcccc}
        \toprule
        \multirow{2}{*}{Hyperparameter} & \multicolumn{2}{c}{Annotation} & \multicolumn{2}{c}{Evaluation} \\
        \cmidrule(lr){2-3} \cmidrule(lr){4-5}
        & ReAct & PnE Executor & ReAct & PnE Executor \\
        \midrule
        \texttt{temperature} & 0.1 & 0.3 & 0.1 & 0.1 \\
        \texttt{top_p} & 0.5 & 0.5 & 0.5 & 0.5 \\
        \texttt{max_context_length} & 40000 & 20000 & 1000000 & 1000000 \\
        \texttt{max_steps} & 50 & 20 & 50 & 50 \\
        \texttt{max_tokens} & 2000 & 2000 & 2000 & 2000 \\
        \bottomrule
    \end{tabular}
    \caption{Decoding hyperparameters for Annotation and ICL Evaluation experiments for ReAct solver and PnE Executor. \texttt{max_context_length} limits the length of the input prompt while \texttt{max_tokens} limits the length of the output. \texttt{max_steps} is the maximum number of steps for the agent to take.}
    \label{tab:hyperparams}
\end{resultstable}

\begingroup
\definecolor{codegreen}{rgb}{0,0.6,0}
\definecolor{codegray}{rgb}{0.5,0.5,0.5}
\definecolor{codepurple}{rgb}{0.58,0,0.82}
\definecolor{bgcolor}{rgb}{1,0.85,1}
\definecolor{envbgcolor}{rgb}{1,1,0.85}
\definecolor{agentbgcolor}{rgb}{0.85,1,1}
\lstdefinestyle{promptstyle}{
    backgroundcolor=\color{bgcolor},
    commentstyle=\color{codegreen},
    keywordstyle=\color{magenta},
    numberstyle=\tiny\color{codegray},
    stringstyle=\color{codepurple},
    basicstyle=\ttfamily\scriptsize,
    breakatwhitespace=false,
    breaklines=true,
    breakautoindent=false,
    breakindent=0pt,
    captionpos=b,
    keepspaces=true,
    numbers=none,
    numbersep=5pt,
    showspaces=false,
    showstringspaces=false,
    showtabs=false,
    tabsize=2,
    frame=single,
}

\lstset{style=promptstyle,belowskip=-0.2\baselineskip}
\begin{figure*}
    \centering
\begin{lstlisting}[backgroundcolor=\color{envbgcolor},firstline=2,lastline=17]
I am your supervisor and you are a super-intelligent AI Assistant whose job is to assist with my day-to-day tasks involving various apps (e.g., amazon.com, gmail, calendar, etc.). To do this, we will take part in a *multi-turn conversation* with a Python REPL environment that will let you interact with the apps using their APIs.

At every step of the conversation, you will need to reason about your current progress on the task and propose the next action in the form of a Python code snippet for the environment to execute. Your response needs to be in the following JSON format {"thought": <thought>, "action": <action>} where <thought> is your reasoning and <action> the Python code snippet (without any enclosing ```).

The environment will then execute your code and respond with the output. The environment will maintain state across multiple interactions for the on-going task so that any variables defined in one step can be used in subsequent steps. Additionally, since you'll be solving a lot of tasks, it is possible that your reasoning on the current step matches with the reasoning at some step of a task you have solved before. When this happens, I will provide you with the relevant snippet of your conversation solving that task. These may be helpful for your next step.

Here are three key APIs that you can use to get more information about apps and APIs:

# To get a list of apps that are available to you:
print(apis.api_docs.show_app_descriptions())

# To get the list of apis under any app listed above, e.g. amazon
print(apis.api_docs.show_api_descriptions(app_name='amazon'))

# To get the full input-output specification of a particular api, e.g. amazon app's login api
print(apis.api_docs.show_api_doc(app_name='amazon', api_name='show_cart'))
\end{lstlisting}
\begin{lstlisting}[backgroundcolor=\color{bgcolor},firstline=19,lastline=19]
<task_trajectory_demonstrations>
\end{lstlisting}
\begin{lstlisting}[backgroundcolor=\color{envbgcolor},firstline=22,lastline=22]
Now, here is the actual task you need to solve using a fresh environment.
\end{lstlisting}
\begin{lstlisting}[backgroundcolor=\color{envbgcolor},firstline=25,lastline=26]
My name is Joyce Weaver. My personal email is joyce-weav@gmail.com and phone number is 3155673041.
Task: Request $13 publicly on Venmo from my friend, Stacy, with a note, "For yesterday's meal".
\end{lstlisting}
\begin{lstlisting}[backgroundcolor=\color{envbgcolor},firstline=29,lastline=47]
Here are some key guidelines that you need to follow:

(1) Make sure to produce a *single* thought and action at every step and correctly format them as {"thought": <thought>, "action": <action>}. In particular,
    - the JSON should be valid with any quotes, newlines, etc., properly escaped.
    - <action> should be just code without any enclosing backticks (```).

(2) Always look at API specifications (using apis.api_docs.show_api_doc) before calling an API.

(3) Remember you can use the variables in your code in subsequent code blocks.

(4) Remember that the email addresses, access tokens and variables (e.g. amazon_password) in the example above are not valid anymore. You will be provided a fresh environment to work with. Note, however, that the APIs remain the same so if it's been shown in an example above, you don't need to look at its specification again.

(5) You can use the "supervisor" app to get information about my accounts and use the "phone" app to get information about friends and family.

(6) Many APIs return items in "pages". Make sure to run through all the pages by looping over `page_index`.

(7) If your action produces output that is too long, the environment will truncate it with '... [HIDDEN FOR BREVITY] ...' replacing the middle part. E.g., this will happen if you print a large number of items returned by a search API. In such cases, you should consider using a more precise query to reduce the number of items returned.

(8) Once you have completed the task, make sure to call apis.supervisor.complete_task(). If the task asked for some information, return it as the answer argument, i.e., call apis.supervisor.complete_task(answer=<answer>). Many tasks do not require an answer, so in those cases, just call apis.supervisor.complete_task() i.e. do not pass any argument.
\end{lstlisting}
    \caption{Task context for the ReAct solver. Each box is a separate message. \texttt{<task trajectory_demonstrations} is the placeholder for any trajectory demonstrations (see Fig. \ref{fig:traj_prompt} for the corresponding template).}
    \label{fig:react_prompt}
\end{figure*}

\begin{figure*}
    \centering
\begin{lstlisting}[backgroundcolor=\color{envbgcolor},firstline=2,lastline=20]
I am your supervisor and you are a super intelligent AI Assistant whose job is to assist with my day-to-day tasks involving various apps (e.g., amazon.com, gmail, calendar, etc.). As the tasks can be complex, I will break them down into a sequence of subtasks that you will need to carry out one at a time. To do this, you will take part in a *multi-turn conversation* with a Python REPL environment that will let you interact with the apps using their APIs.

At every step of the conversation, you will need to reason about your current progress on the subtask and propose the next action in the form of a Python code snippet for the environment to execute. Your response needs to be in the following JSON format {"thought": <thought>, "action": <action>} where <thought> is your reasoning and <action> the Python code snippet (without any enclosing ```).

Finally, when you have completed the subtask, you will need to say FINISH as action in a separate response in the same format, i.e.,
{"thought": <thought>, "action": FINISH}

The environment will then execute your code and respond with the output. The environment will maintain state across multiple interactions for the on-going task so that any variables defined in one step can be used in subsequent steps. Additionally, when your reasoning for the current step matches with a task you have solved previously, I will also provide you with the relevant portion of the conversation that solved that task to help you on subsequent steps for the current task.

Here are three key APIs that you can use to get more information about apps and APIs:

# To get a list of apps that are available to you:
print(apis.api_docs.show_app_descriptions())

# To get the list of apis under any app listed above, e.g. amazon
print(apis.api_docs.show_api_descriptions(app_name='amazon'))

# To get the full input-output specification of a particular api, e.g. amazon app's login api
print(apis.api_docs.show_api_doc(app_name='amazon', api_name='show_cart'))
\end{lstlisting}
\begin{lstlisting}[backgroundcolor=\color{bgcolor},firstline=22,lastline=22]
<subtask_trajectory_demonstrations>
\end{lstlisting}
\begin{lstlisting}[backgroundcolor=\color{envbgcolor},firstline=25,lastline=25]
Now, here is the actual task you need to solve using a fresh environment.
\end{lstlisting}
\begin{lstlisting}[backgroundcolor=\color{envbgcolor},firstline=28,lastline=38]
My name is Joyce Weaver. My personal email is joyce-weav@gmail.com and phone number is 3155673041.
Task: Request $13 publicly on Venmo from my friend, Stacy, with a note, "For yesterday's meal".

The above task can be decomposed into the following subtasks that need to be carried out one by one

1. Login to Venmo and save access token in `venmo_access_token` variable.
2. Use the `venmo_access_token` to search for Stacy in my Venmo contacts and save her user ID in `stacy_user_id`.
3. Use the `venmo_access_token` to create a payment request to `stacy_user_id` for $13 with the note 'For yesterday's meal'. Ensure the request is set to public.
4. Complete task.

Let's start by solving Subtask 1: Login to Venmo and save access token in `venmo_access_token` variable.
\end{lstlisting}
\begin{lstlisting}[backgroundcolor=\color{envbgcolor},firstline=41,lastline=61]
Here are some key guidelines that you need to follow:

(1) Do not worry about the entire task. Focus ONLY on the current subtask and carry it out carefully and correctly.

(2) Make sure to produce a *single* thought and action at every step and correctly format them as {"thought": <thought>, "action": <action>}. In particular,
    - the JSON should be valid with any quotes, newlines, etc., properly escaped.
    - <action> should be just code without any enclosing backticks (```).

(3) When finished with the current subtask, make sure to say FINISH in a SEPARATE response in the format described above. Also, if it's the last subtask, make sure to call `apis.supervisor.complete_task()` with the answer, if any, before finishing.

(4) Always look at API specifications (using apis.api_docs.show_api_doc) before calling an API.

(5) Remember you can use the variables in your code in subsequent code blocks.

(6) Remember that the email addresses, access tokens and variables (e.g. amazon_password) in the example above are not valid anymore. You will be provided a fresh environment to work with. Note, however, that the APIs remain the same so if it's been shown in an example above, you don't need to look at its specification again.

(7) You can use the "supervisor" app to get information about my accounts and use the "phone" app to get information about friends and family.

(8) Many APIs return items in "pages". Make sure to run through all the pages by looping over `page_index`.

(9) If your action produces output that is too long, the environment will truncate it with '... [HIDDEN FOR BREVITY] ...' replacing the middle part. E.g., this will happen if you print a large number of items returned by a search API. In such cases, you should consider using a more precise query to reduce the number of items returned.
\end{lstlisting}
    \caption{Task context for the PnE executor. Each box is a separate message. \texttt{<subtask trajectory_demonstrations} is the placeholder for any trajectory demonstrations (see Fig. \ref{fig:subtraj_prompt} for the corresponding template).}
    \label{fig:executor_prompt1}
\end{figure*}

\begin{figure*}
    \centering
\begin{lstlisting}[backgroundcolor=\color{envbgcolor}]
I am your supervisor and you are a super intelligent AI Assistant whose job is to autonomously perform my day-to-day tasks involving various apps (e.g., spotify, simple_note, etc.).

To do this, you will need to interact with apps using their associated APIs on my behalf.
Here are the apps:
{ app_descriptions }

To help with this, I will provide you with an executor model that will take care of interacting with the APIs. Your job is to produce a plan on how to solve the task given access to this executor. You should respond in the following JSON format:


{
    "thought": <thought>,
    "plan": [
        "<subtask_1>",
        "<subtask_2>",
        ...
        "<subtask_n>"
    ]
}


Here, <thought> is a brief description of how you plan to solve the task and <subtask_i> is a description of the ith subtask in your plan.

**Key instructions**:
(1) Make sure to respond in the JSON format provided above. Don't enclose your response with ```.

(2) Make sure to define all the relevant variables.

(3) Each subtask in the plan should be clear and complete so that it can be executed independently. E.g. don't use references such as  "previous list" or "repeat subtask 1".

(4) When unsure, use more subtasks in the plan rather than fewer subtasks.

(5) The final subtask should be "Complete task." unless the task requires an answer, in which case, it should be "Complete task with answer: <answer>" where <answer> is the answer to the task.

\end{lstlisting}
    \caption{Prompt for the PnE planner. It is followed by a few demonstrative task-plan pairs and the test task.}
    \label{fig:planner_prompt}
\end{figure*}

\begin{figure*}
    \centering
\begin{lstlisting}[backgroundcolor=\color{envbgcolor},firstline=2,lastline=2]
Here is an example of a task and how you can interact with the environment to solve it
\end{lstlisting}
\begin{lstlisting}[backgroundcolor=\color{envbgcolor},firstline=5,lastline=6]
My name is Joyce Weaver. My personal email is joyce-weav@gmail.com and phone number is 3155673041.
Task: How many playlists do I have in Spotify?
\end{lstlisting}
\begin{lstlisting}[backgroundcolor=\color{agentbgcolor},firstline=9,lastline=12]
Lets first find which APIs are available to use in Spotify.
```python
print(apis.api_docs.show_api_descriptions(app_name='spotify'))
```
\end{lstlisting}
\begin{lstlisting}[backgroundcolor=\color{envbgcolor},firstline=15,lastline=29]
Output:
```
[
 ...
 {
  "name": "login",
  "description": "Login to your account."
 },
 {
  "name": "logout",
  "description": "Logout from your account."
 },
 ...
]
```
\end{lstlisting}
    $\vdots$
\begin{lstlisting}[backgroundcolor=\color{agentbgcolor},firstline=32,lastline=35]
Now that the task is completed, I need to mark the task as complete and return the number of playlists found.
```python
apis.supervisor.complete_task(answer=num_playlists)
```
\end{lstlisting}
\begin{lstlisting}[backgroundcolor=\color{envbgcolor},firstline=38,lastline=41]
Output:
```
Marked the active task complete.
```
\end{lstlisting}
    \caption{Template used for trajectory demonstrations. Yellow boxes are user messages or environment responses, while blue boxes are agent messages (originally in JSON format).}
    \label{fig:traj_prompt}
\end{figure*}

\begin{figure*}
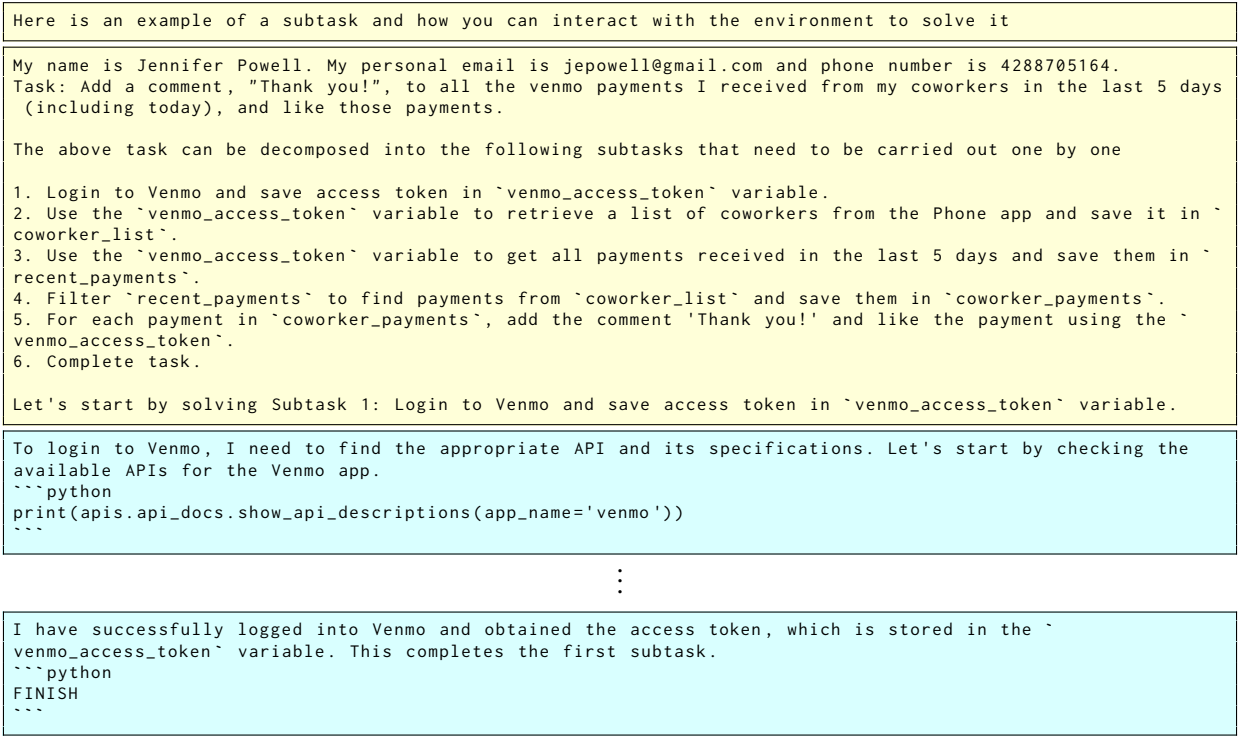

    \centering
\begin{lstlisting}[backgroundcolor=\color{envbgcolor},firstline=2,lastline=2]
Here is an example of a subtask and how you can interact with the environment to solve it
\end{lstlisting}
\begin{lstlisting}[backgroundcolor=\color{envbgcolor},firstline=5,lastline=17]
My name is Jennifer Powell. My personal email is jepowell@gmail.com and phone number is 4288705164.
Task: Add a comment, "Thank you!", to all the venmo payments I received from my coworkers in the last 5 days (including today), and like those payments.

The above task can be decomposed into the following subtasks that need to be carried out one by one

1. Login to Venmo and save access token in `venmo_access_token` variable.
2. Use the `venmo_access_token` variable to retrieve a list of coworkers from the Phone app and save it in `coworker_list`.
3. Use the `venmo_access_token` variable to get all payments received in the last 5 days and save them in `recent_payments`.
4. Filter `recent_payments` to find payments from `coworker_list` and save them in `coworker_payments`.
5. For each payment in `coworker_payments`, add the comment 'Thank you!' and like the payment using the `venmo_access_token`.
6. Complete task.

Let's start by solving Subtask 1: Login to Venmo and save access token in `venmo_access_token` variable.
\end{lstlisting}
\begin{lstlisting}[backgroundcolor=\color{agentbgcolor},firstline=20,lastline=23]
To login to Venmo, I need to find the appropriate API and its specifications. Let's start by checking the available APIs for the Venmo app.
```python
print(apis.api_docs.show_api_descriptions(app_name='venmo'))
```
\end{lstlisting}
    $\vdots$
\begin{lstlisting}[backgroundcolor=\color{agentbgcolor},firstline=26,lastline=29]
I have successfully logged into Venmo and obtained the access token, which is stored in the `venmo_access_token` variable. This completes the first subtask.
```python
FINISH
```
\end{lstlisting}
    \caption{Template used for subtask trajectory demonstrations. Yellow boxes are user messages or environment responses, while blue boxes are agent messages (originally in JSON format). The main difference with the task trajectory demonstrations (Fig. \ref{fig:traj_prompt}) is that the agent is additionally provided the plan and a summary of code used to solve previous subtasks.}
    \label{fig:subtraj_prompt}
\end{figure*}

\begin{figure*}
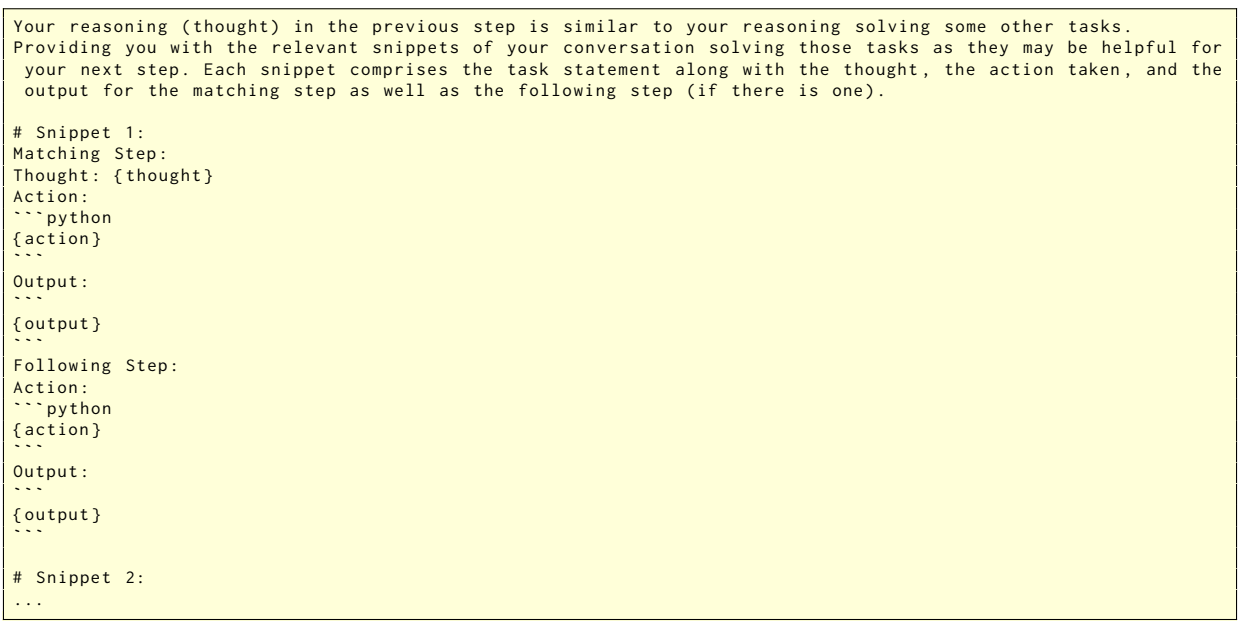

    \centering
\begin{lstlisting}[backgroundcolor=\color{envbgcolor}]
Your reasoning (thought) in the previous step is similar to your reasoning solving some other tasks. Providing you with the relevant snippets of your conversation solving those tasks as they may be helpful for your next step. Each snippet comprises the task statement along with the thought, the action taken, and the output for the matching step as well as the following step (if there is one).

# Snippet 1:
Matching Step:
Thought: {thought}
Action:
```python
{action}
```
Output:
```
{output}
```
Following Step:
Action:
```python
{action}
```
Output:
```
{output}
```

# Snippet 2:
...
\end{lstlisting}
    \caption{Template used for showing snippet demonstrations. This is a single message appended after to all the messages.}
    \label{fig:snippet_prompt}
\end{figure*}
\endgroup

\section{Trained Baselines}
\label{app:trained}

We compare with the following training-based approaches baselines from \citet{loop}:
\begin{itemize}
    \item \textbf{Ground truth supervised fine-tuning (SFT-GT)}. SFT on ReAct-style transformation of gold solutions.

    \item \textbf{Rejection sampling fine-tuning (RFT) \citep{rft}}. Collects rollouts generated with the base model and finetunes on successful ones.
    \item \textbf{Expert iteration (EI) \citep{ei}}. Runs multiple smaller iterations of RFT using the current best model.
    \item \textbf{Direct Preference Optimization + MCTS (DPO-MCTS) \citep{dpo}}. Collects preference pairs into a replay buffer using Monte-Carlo Tree Search.
    \item \textbf{Proximal Policy Optimization (PPO) \citep{ppo}}. PPO with a learned advantage estimate.
    \item \textbf{REINFORCE leave-one-out (RLOO) \citep{rloo}}. On-policy trajectory-level REINFORCE with leave-one-out advantage estimate.
    \item \textbf{Group relative policy optimization (GRPO) \citep{grpo}}. On-policy PPO with normalized leave-one-out advantage estimate.
    \item \textbf{Leave-one-out PPO (LOOP) \citep{loop}}. Off-policy PPO with unnormalized leave-one-out advantage estimate.
\end{itemize}

\end{document}